%% file: main_arxiv.tex
\documentclass[10pt,twocolumn,letterpaper]{article}

\usepackage[pagenumbers]{cvpr}

\usepackage{subcaption}
\usepackage{colortbl}
\definecolor{light-light-gray}{gray}{0.92} 
\usepackage{listings}
\usepackage{tcolorbox}

\definecolor{cvprblue}{rgb}{0.21,0.49,0.74}
\usepackage[pagebackref,breaklinks,colorlinks,allcolors=cvprblue]{hyperref}
\usepackage{algorithm}
\usepackage{algorithmic}

\title{Automated Generation of Challenging Multiple-Choice Questions for \\ Vision Language Model Evaluation}

\author{
\textbf{Yuhui Zhang$^{1}$\thanks{Equal contribution. $^\dagger$Equal advising. Correspondence to: \tt{\{yuhuiz,suyc,ludwigsc,syyeung\}@stanford.edu}. Project page: \url{https://yuhui-zh15.github.io/AutoConverter-Website/}.} \quad Yuchang Su$^{2*}$ \quad Yiming Liu$^{2}$ \quad Xiaohan Wang$^{1}$ \quad James Burgess$^{1}$} \\ 
\textbf{Elaine Sui$^{1}$ \quad  Chenyu Wang$^{3}$ \quad Josiah Aklilu$^{1}$ \quad Alejandro Lozano$^{1}$ \quad Anjiang Wei$^{1}$} \\ 
\textbf{Ludwig Schmidt$^{1\dagger}$ \quad Serena Yeung-Levy$^{1\dagger}$} \\
$^{1}$Stanford University \quad $^{2}$Tsinghua University \quad $^{3}$MIT
}

\begin{document}
\maketitle

\input{sections/0_abstract}
\input{sections/1_intro}

\input{sections/5_related}
\input{sections/2_analysis}
\input{sections/3_method}
\input{sections/4_benchmark}
\input{sections/6_conclusion}

{
    \small
    \bibliographystyle{ieeenat_fullname}
    \bibliography{main}
}

\newpage
\thispagestyle{empty}
\null
\newpage

\appendix
\input{appendix_arxiv}

\end{document}

%% file: sections/0_abstract.tex
\begin{abstract}

The rapid development of vision language models (VLMs) demands rigorous and reliable evaluation. However, current visual question answering (VQA) benchmarks often depend on open-ended questions, making accurate evaluation difficult due to the variability in natural language responses. To address this, we introduce AutoConverter, an agentic framework that automatically converts these open-ended questions into multiple-choice format, enabling objective evaluation while reducing the costly multiple-choice question creation process. Our experiments demonstrate that AutoConverter can generate correct and challenging multiple-choice questions, with VLMs demonstrating consistently similar or lower accuracy on these questions compared to human-created ones. Using AutoConverter, we construct VMCBench, a benchmark created by transforming 20 existing VQA datasets into a unified multiple-choice format, totaling 9,018 questions. We comprehensively evaluate 33 state-of-the-art VLMs on VMCBench, setting a new standard for scalable, consistent, and reproducible VLM evaluation.

\end{abstract}
\vspace{-1em}

%% file: sections/1_intro.tex
\section{Introduction}

\begin{figure*}
    \centering
    \includegraphics[width=\linewidth]{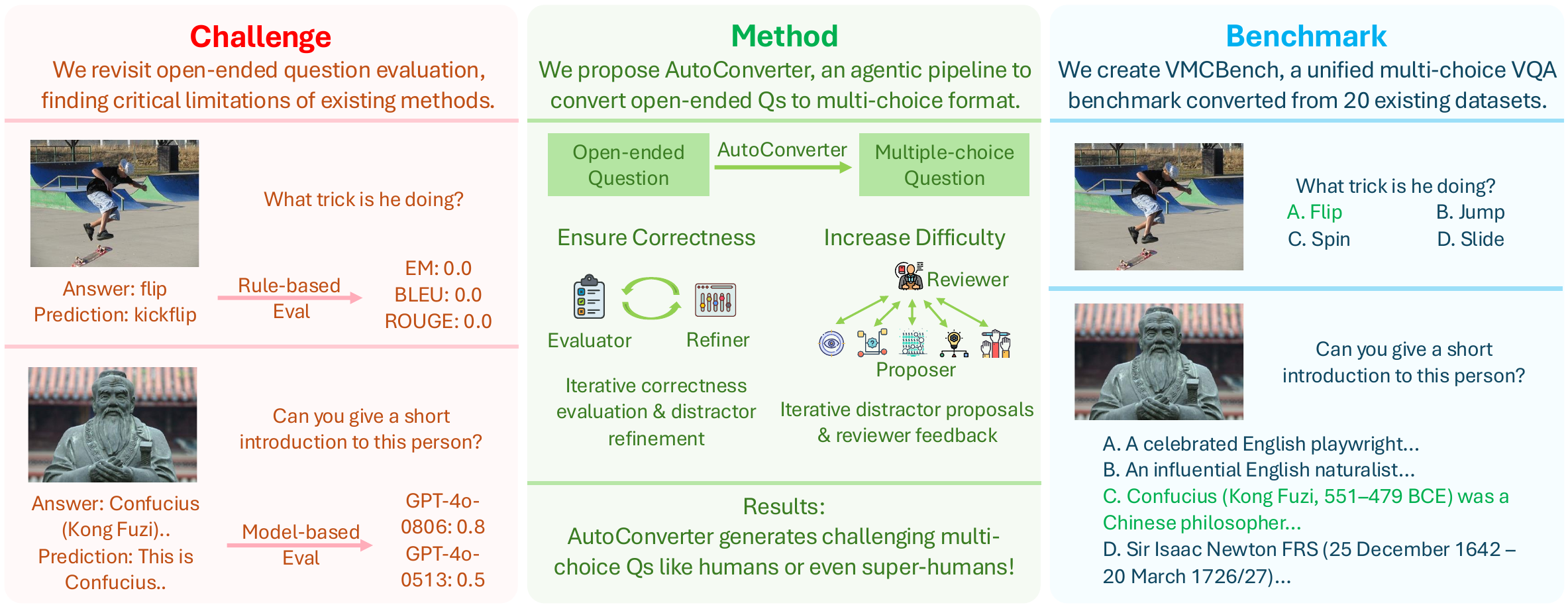}
    \vspace{-2em}
    \caption{\textbf{Overview.} 
    \textit{(Left)} We analyze existing open-ended VQA evaluation metrics, underscoring their limitations in providing accurate and reproducible assessments.
    \textit{(Middle)} We introduce \textit{AutoConverter}, a multi-agent system that automatically converts open-ended questions into multiple-choice format, enabling objective assessment while reducing the costly question creation process. 
    \textit{(Right)} Using \textit{AutoConverter}, we convert and refine 20 existing VQA datasets into a unified multiple-choice benchmark to support future VLM research.
    }
    \vspace{-1em}
    \label{fig:pull}
\end{figure*}

With the rapid advancements in vision language models (VLMs)~\cite{gpt4,claude,team2023gemini,llava1.5}, rigorous and reliable evaluation methods are critical to gauge model performance and guide further research. Visual question answering (VQA) has emerged as a standard evaluation protocol, typically structured with open-ended or multiple-choice questions. Open-ended questions~\cite{vqav2,okvqa,mmvet} allow models to generate free-form, natural language responses, while multiple-choice questions~\cite{aokvqa,scienceqa,mmmu} ask models to choose the answer from predefined options. Most existing VQA benchmarks are in open-ended format~\cite{duan2024vlmevalkit,lee2024vhelm} because of the complexity and extensive resources required to design high-quality multiple-choice questions.

We revisit current evaluation strategies for open-ended questions, focusing on the challenge of measuring semantic similarity between model-generated and ground-truth answers (\S\ref{sec:challenge}).
Two main evaluation methods exist: rule-based, which evaluates word or phrase level overlap, and model-based, which uses large language models (e.g., GPT-4o) for semantic matching.
While rule-based evaluation is computationally efficient, our experiments on the VQAv2 dataset~\cite{vqav2} demonstrate that it fails to capture semantic nuances and formatting differences, yielding a poor correlation (0.09) with true VLM performance. In contrast, model-based metrics, though more semantically accurate, are costly and unstable --- updates in model versions (e.g., GPT-4o from 0513 to 0806) constantly increase scores by 6\% on the MMVet dataset~\cite{mmvet}, making results incomparable and raising future reproducibility issues.

To mitigate these issues, we propose an alternative for VLM evaluation by converting open-ended VQA questions into multiple-choice format. This shift allows for more objective and reproducible assessment by providing defined options and simplifying answer validation. Nonetheless, creating multiple-choice questions is inherently complex, particularly in generating distractor options that are plausible yet challenging given the correct answer --- a process traditionally requiring extensive human expertise and effort to simulate different errors students could make.

We introduce \textit{AutoConverter} (\S\ref{sec:method}), a novel multi-agent system powered by GPT-4o, designed to automatically convert open-ended questions into challenging yet correct multiple-choice questions. To enhance difficulty, specialized distractor proposer agents collaborate with a reviewer agent in an iterative manner to generate a large pool of challenging distractors focused on common error types, including conceptual misunderstandings, visual misinterpretations, and reasoning errors. A selector agent then chooses the most challenging distractors. To ensure correctness, a question correctness evaluator agent assesses the similarity between distractors and the correct answer, providing feedback that allows a refiner agent to adjust distractors to improve question correctness. This agentic pipeline ensures that the generated multiple-choice questions are both correct and challenging, delivering strong discriminative power for VLM evaluation.

Our experiments validate \textit{AutoConverter}'s ability to produce correct and challenging multiple-choice questions. When applied to established multiple-choice VQA datasets such as MMMU~\cite{mmmu}, MathVista~\cite{mathvista}, and AI2D~\cite{ai2d}, \textit{AutoConverter} generates distractors that match or even surpass the difficulty of human-crafted alternatives, as evidenced by comparable or lower accuracy scores across various VLMs, with only 3\% of questions with the highest correctness score marked incorrect by human annotators. A detailed ablation study further demonstrates that each component of \textit{AutoConverter} significantly enhances the correctness and difficulty of questions compared to naive conversion methods. These results indicate \textit{AutoConverter}'s broad applicability beyond converting open-ended questions to multiple-choice format, such as refining existing multiple-choice datasets to increase difficulty or generating challenging questions for students in an educational context.

Building on \textit{AutoConverter}, we introduce \textit{\textit{VMCBench}}, a benchmark comprising 9,018 multiple-choice VQA questions compiled from 20 existing datasets (\S\ref{sec:benchmark}). These original datasets, which contain either open-ended or multiple-choice questions, have been converted or refined using \textit{AutoConverter}. Any questions flagged as uncertain by the correctness evaluator were manually verified to ensure quality and accuracy. \textit{VMCBench} provides a standardized benchmark to evaluate diverse model capabilities across various question types. We evaluated 33 state-of-the-art VLMs on \textit{VMCBench}, establishing a new standard for scalable, consistent, and reproducible vision language model evaluation. Our contributions are summarized in Figure~\ref{fig:pull}.

%% file: sections/5_related.tex
\section{Related Works}

\begin{figure*}[!tb]
\centering
\includegraphics[width=\linewidth]{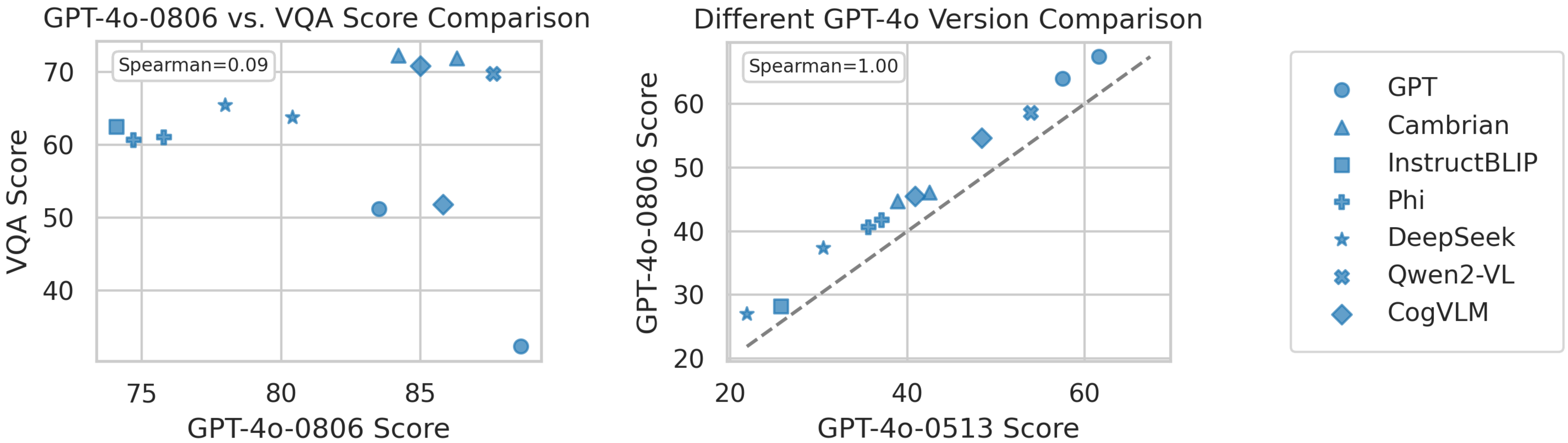}
\vspace{-2em}
\caption{\textbf{Challenges in evaluating open-ended questions.} \textit{(Left)} Rule-based metrics significantly underestimate model performance and penalize models that do not strictly follow the expected format. \textit{(Right)} Model-based evaluations using two different versions of GPT yield substantially different scores, making comparisons inconsistent and raising reproducibility issues. The repeated points represent different model sizes within the same model family (e.g., GPT-4o/GPT-4o-Mini for 2 points).
}
\label{fig:eval_failure}
\vspace{-1em}
\end{figure*}

\textbf{Vision language model evaluation.}
With the rapid development of vision language models (VLMs)~\cite{gpt4,claude,team2023gemini,llava1.5}, numerous benchmarks have emerged to assess various capabilities of VLMs, including general understanding~\cite{vqav2,okvqa,mmvet,mmstar,aokvqa,vizwiz,seedbench}, reasoning~\cite{mathvista,mmmu,mathvision,scienceqa,realworldqa,gqa}, OCR~\cite{textvqa,ocrvqa}, and document and chart comprehension~\cite{docvqa,ai2d,chartqa,infovqa,tablevqa}. These datasets typically use either open-ended or multiple-choice formats. Open-ended questions are easier to create but challenging to evaluate accurately, while multiple-choice questions simplify evaluation but demand greater effort in their creation. Previous studies have found limitations of open-ended evaluation for VLMs~\cite{chenmllm,manas2024improving,laurenccon2024building}, often focusing on case studies and comparisons with human performance. Their proposed solutions are model-based evaluation methods~\cite{zheng2023judging,dubois2023alpacafarm}.
In this work, we systematically analyze various VLMs on two widely-used open-ended VQA datasets in zero-shot evaluation settings, quantitatively revealing significant challenges in VLM evaluation for open-ended questions. We demonstrate how rule-based evaluation poorly correlates with actual model rankings in zero-shot scenarios. Additionally, we highlight the substantial limitations of model-based evaluation: model-based evaluations are inconsistent across versions, which leads to serious reproducibility issues. These issues seem intrinsic and difficult to resolve. Consequently, we propose to convert open-ended questions to multiple-choice format in order to streamline reliable VLM evaluation.

\textbf{Converting open-ended to multiple-choice.}
Converting open-ended questions to multiple-choice requires generating distractors that are both challenging and accurate—a task that traditionally demands extensive human expertise and has limited the construction of multiple-choice benchmarks. Some approaches attempt to use knowledge graphs~\cite{yu2024enhancing}, retrieval~\cite{luo2024chain}, reinforcement learning~\cite{lu2022good}, and supervised learning~\cite{ding2024can} to automatically produce distractors that resemble correct answers. However, these methods require external resources and often result in relatively simple distractors. Recent advances in large language models (LLMs) have inspired attempts to automate distractor generation to support human creators, yet these approaches still necessitate significant human input to ensure question quality~\cite{mmlupro,mmmupro}. In our work, we systematically define two essential criteria—correctness and difficulty—for effective multiple-choice questions, and develop an agentic pipeline~\cite{park2023generative,shinn2024reflexion,liuagentbench} to automate their creation. Our approach generates multiple-choice questions with distractors that are both challenging and correct, achieving quality comparable to or even surpassing human-crafted questions while significantly reducing the need for human effort. This tool has broad utility: it can convert open-ended questions to multiple-choice for VLM evaluation, refine poorly constructed questions, and generate questions beyond VLM evaluation, such as educational assessments.

%% file: sections/2_analysis.tex
\section{Open-Ended Question Evaluation Challenge}
\label{sec:challenge}

\input{tables/autoconverter}

In this section, we discuss the challenges of evaluating vision language models (VLMs) using open-ended questions. The primary issue lies in accurately and robustly measuring the semantic similarity between model-generated answers and ground-truth answers, a long-standing challenge in natural language processing~\cite{fabbri2021summeval,zhang2024benchmarking,zheng2023judging}.

\subsection{Rule-based Evaluation}

\textbf{Background.} When ground-truth answers are short, rule-based evaluation metrics such as exact match or quasi-exact match (e.g., BLEU~\cite{bleu}, ROUGE~\cite{rouge}) are typically used to measure word-level overlap between the ground truth and model predictions. However, these methods are limited in accounting for semantic variations, such as synonyms, paraphrasing, or formatting differences. This limitation is even more pronounced when evaluating current VLMs in zero-shot settings, where models are instruction-tuned to generate slightly more detailed and lengthier responses for clarity.

\textbf{Experiment.} To examine the limitations of these evaluation metrics, we tested 12 state-of-the-art VLMs on the widely-used VQAv2 dataset~\cite{vqav2}. In a zero-shot setting, we prompted the models with, “Please try to answer the question with short words or phrases if possible.” and used VQAv2’s official rule-based evaluation code. We then conducted a detailed analysis of the cases where GPT-4o did not receive full marks, using human judgment to correct scores where the rule-based evaluation failed. Based on these human annotations, we developed a model-based evaluation method using GPT-4o with a well-crafted prompt to assess the semantic match between model outputs and ground-truth answers. Our model-based metric achieved a 0.95 agreement with human evaluations on VQAv2, making it a reliable proxy for human judgment.

\textbf{Findings.} Analysis of GPT-4o’s errors revealed that approximately 66\% of the misclassifications were due to evaluation failures rather than the model’s inability to answer the questions. With human corrections, GPT-4o achieved an accuracy of 88\% on VQAv2, significantly higher than the 32\% accuracy reported by rule-based metrics. A comparison between rule-based and model-based evaluations (i.e., human-proxy evaluation) showed a Spearman correlation of only 0.09, as seen in Figure~\ref{fig:eval_failure} (left), demonstrating that rule-based metrics provide nearly random and unreliable scores, failing to distinguish between state-of-the-art VLMs.

\subsection{Model-based Evaluation}

\textbf{Background.} When ground-truth answers are longer, reflecting more realistic use cases for VLMs, rule-based metrics become even less effective at capturing semantic equivalence between predictions and answers. Consequently, the community increasingly relies on model-based evaluation~\cite{dubois2023alpacafarm,zheng2023judging}, which leverages advances in language models like GPT-4o with carefully designed prompts to generate similarity scores between predictions and answers. Although these similarity scores correlate well with human evaluations, changes in model versions can lead to significant shifts in evaluation results.

\textbf{Experiment.} To evaluate the stability and robustness of model-based metrics, we tested 12 state-of-the-art VLMs on the MMVet dataset~\cite{mmvet}, where answer length is 63 words on average. Using two versions of GPT-4o (GPT-4o-0806 and GPT-4o-0513) as evaluators, we kept all other variables constant, such as model responses and evaluation prompts.

\textbf{Findings.} While model-based evaluation is reliable in ranking models comparatively, it is costly and introduces notable shifts in absolute scores. As shown in Figure~\ref{fig:eval_failure} (right), we observed a perfect 1.0 correlation in VLM rankings between the two GPT-4o versions, underscoring the model-based evaluation’s strong alignment with human judgment. However, the absolute scores from GPT-4o-0806 were consistently 6\% higher than those from GPT-4o-0513. Further analysis indicated that GPT-4o-0806 tended to assign a score of 1.0 instead of 0.9 for nearly-correct responses, while GPT-4o-0513 showed the opposite tendency. These differences in absolute values hinder the comparability of results across different model versions, significantly impacting the reproducibility of research findings, especially when older versions are deprecated. 

%% file: tables/autoconverter.tex
\captionsetup[sub]{font=small}
\begin{figure*}[!tb]
  \centering
  \begin{subfigure}[b]{0.49\textwidth}
    \centering
    \includegraphics[width=\linewidth]{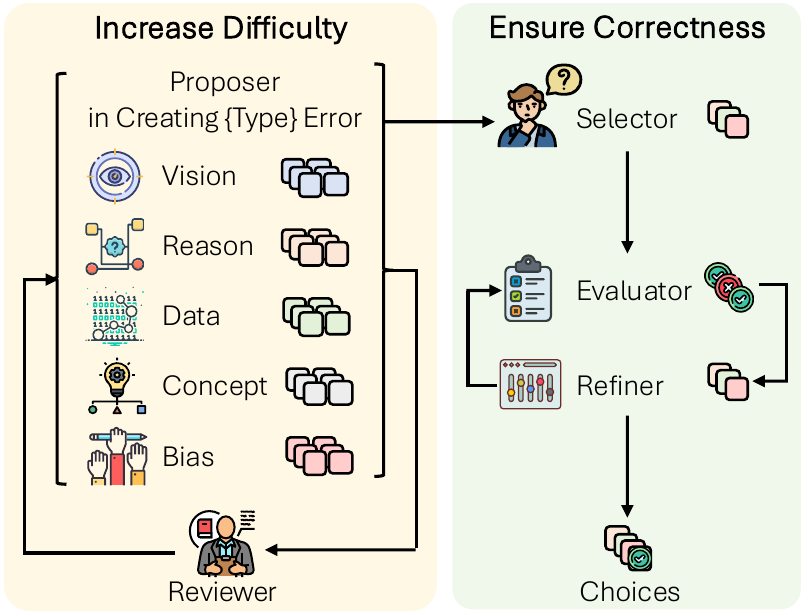}
    \caption{\textit{AutoConverter} framework.}
    \label{fig:method}
  \end{subfigure}
  \hfill
  \begin{subfigure}[b]{0.49\textwidth}
    \centering
    \small
    \rowcolors{2}{white}{light-light-gray}
    \begin{tabular}{lccccccc}
    \toprule
       Method  & $C$ ($\uparrow$) & $M_1$ & $M_2$ & $M_3$ & $M_4$ & Avg ($\downarrow$) \\
    \midrule
        Human & 4.59 & 33.4 & 35.2 & 46.0 & 52.4 & 41.8 \\
    \midrule
        Naive & 4.59 & 34.2 & 40.6 & 50.2 & 59.0 & 46.0  \\    
    \midrule
        w/o Concept & 4.66 & 33.6 & 37.8 & 46.4 & 57.4 & 43.8 \\ 
        w/o Reason & 4.69 & 30.0 & 40.2 & 47.8 & 56.2 & 43.5 \\ 
        w/o Vision & 4.68 & 29.0 & 39.2 & 45.8 & 57.2 & 42.8 \\ 
        w/o Data & 4.66 & 29.4 & 36.6 & 47.0 & 57.8 & 42.7 \\ 
        w/o Bias & 4.65 & 30.6 & 37.6 & 48.6 & 60.2 & 44.2 \\
        w/o Reviewer & 4.64 & 30.2 & 38.2 & 45.4 & 57.2 & 42.7 \\
        w/o Refiner & 4.28 & 25.0 & 34.2 & 42.0 & 50.0 & 37.8 \\
    \midrule
    \textit{AutoConverter} & 4.69 & 27.8 & 37.2 & 44.2 & 53.6 & 40.7 \\
    \bottomrule
    \end{tabular}
    \caption{Ablation of \textit{AutoConverter} on MMMU. $C$ is correctness (higher is better), Avg is average model performance (lower is better). $M_1$ is PaliGemma-3B, $M_2$ is LLaVA-1.5-7B, $M_3$ is Phi-3.5-Vision, $M_4$ is Qwen2-VL-7B.}
    \label{tab:ablation}
  \end{subfigure}
  \vspace{-0.5em}
  \caption{\textbf{\textit{AutoConverter} framework and results.} \textit{(Left)} \textit{AutoConverter} is a multi-agent framework with two key steps: increasing difficulty and ensuring the correctness of the converted question. \textit{(Right)} We perform an ablation study on \textit{AutoConverter} and find that each component is crucial for enhancing question correctness and achieving the desired level of difficulty.}
  \vspace{-1em}
  \label{fig:combined}
\end{figure*}

%% file: sections/3_method.tex
\section{\textit{AutoConverter}: An Agentic Pipeline Generating Challenging Multi-Choice Questions}
\label{sec:method}

\begin{figure*}[!tb]
    \centering
    \includegraphics[width=\linewidth]{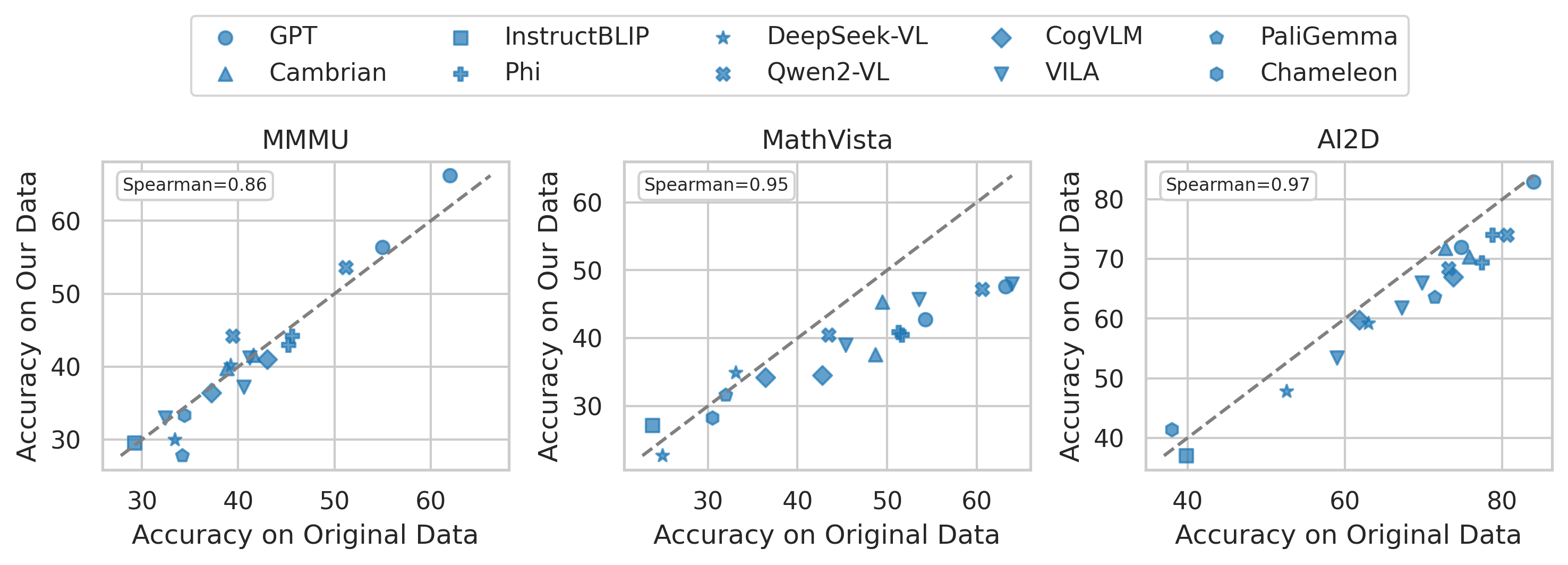}
    \vspace{-2em}
    \caption{\textbf{\textit{AutoConverter} generates challenging multiple-choice questions.} Using \textit{AutoConverter}, we generated distractors for questions and answers from three existing multiple-choice datasets: MMMU, MathVista, and AI2D, and compared them with original human-created distractors. We evaluated various VLMs on both the \textit{AutoConverter}-generated and the original questions, finding that VLMs consistently achieved similar or even lower accuracy on the \textit{AutoConverter}-generated questions compared to the original ones.
    }
    \vspace{-1em}
    \label{fig:difficulty}
\end{figure*}

Given the challenge of evaluating open-ended questions for vision language models (VLMs) detailed in \S\ref{sec:challenge}, how can we mitigate these issues? We propose to convert open-ended questions into a multiple-choice format, capitalizing on the simplicity and objectivity of evaluating multiple-choice questions. However, traditionally, creating multiple-choice questions, especially reasonable yet challenging distractor options, requires substantial human expertise and effort. In this section, we introduce \textit{AutoConverter}, an agentic pipeline that automatically generates high-quality multiple-choice questions from open-ended ones.

\subsection{Problem Formulation and Key Desiderata}

We define the multiple-choice question conversion process as follows: given an image, a question, and the correct answer $(v, q, a) \in \mathcal{X} \times \mathcal{Y} \times \mathcal{Y}$, where $\mathcal{X}$ and $\mathcal{Y}$ represent sets of images and texts respectively, we aim to generate a set of distractor choices $\mathcal{C}_d =\{c_i \in \mathcal{Y} \mid i \in [N]\}$, where $N$ is the number of distractors. We define the candidate choice set as $\mathcal{C} = \mathcal{C}_d \cup \{a\}$, forming the final multiple-choice question as $(v, q, \mathcal{C})$. We choose $N=3$ in our work because 4-choice is the most common configuration for multiple-choice questions and minimizes the risk of option selection bias for language models~\cite{zhao2021calibrate}.

Two key desiderata are crucial for the multiple-choice conversion process. The first is \textbf{correctness}, ensuring that the multiple-choice question is valid with only one correct answer. The second is \textbf{difficulty}, ensuring that the question cannot be answered trivially and possesses sufficient discriminative power for test takers. 
We leverage recent advances in language model agent research, particularly in role-playing~\cite{park2023generative} and self-reflection techniques~\cite{shinn2024reflexion}, to accomplish these goals.

\subsection{Ensuring Correctness}

Correctness is essential when converting open-ended questions to a multiple-choice format, defined as having exactly one correct answer. Since our conversion process retains the original open-ended question stem and its answer, we only need to ensure that the other distractors are incorrect given the question and its correct answer. To achieve this, we employ a multi-agent framework to rigorously evaluate distractor options, further enhancing the correctness of the generated multiple-choice questions through iterative refinement.

\textbf{Evaluating correctness.} We use GPT-4o as a correctness checker, which evaluates each distractor's similarity to the correct answer and assigns a 5-point Likert score for the overall correctness of the question. A score of 5 indicates strong confidence in having a single correct answer, while a score of 1 suggests ambiguity or the presence of multiple correct answers. The evaluator was developed based on a small-scale study of correct and incorrect questions. To validate the effectiveness of this evaluator, we conducted large-scale human annotations on 2,400 questions to assess their correctness (details in \S\ref{sec:benchmark}). Our results indicate that 51\%, 51\%, 63\%, 84\%, and 95\% of questions are deemed correct when assigned correctness scores of 1, 2, 3, 4, and 5, respectively, by our evaluator. Thus, we can use this correctness evaluator as a judge to assess question correctness and refine questions with low correctness scores.

\textbf{Refining questions to enhance correctness.} Since the correctness evaluator can accurately flag problematic questions, we refine questions with low correctness scores in an iterative manner to enhance their accuracy. Specifically, when the correctness score of a generated question falls below a threshold of 4, we use GPT-4o as a refiner to adjust distractors based on feedback from the correctness evaluator. This refinement continues until the evaluator’s correctness score meets the required threshold or until a maximum of three refinement rounds is reached. 

The entire process is shown in Figure~\ref{fig:method} (right). Our ablation studies demonstrate that this process plays a significant role in enhancing the correctness of the questions.

\begin{figure*}[!tb]
    \centering
    \includegraphics[width=\linewidth]{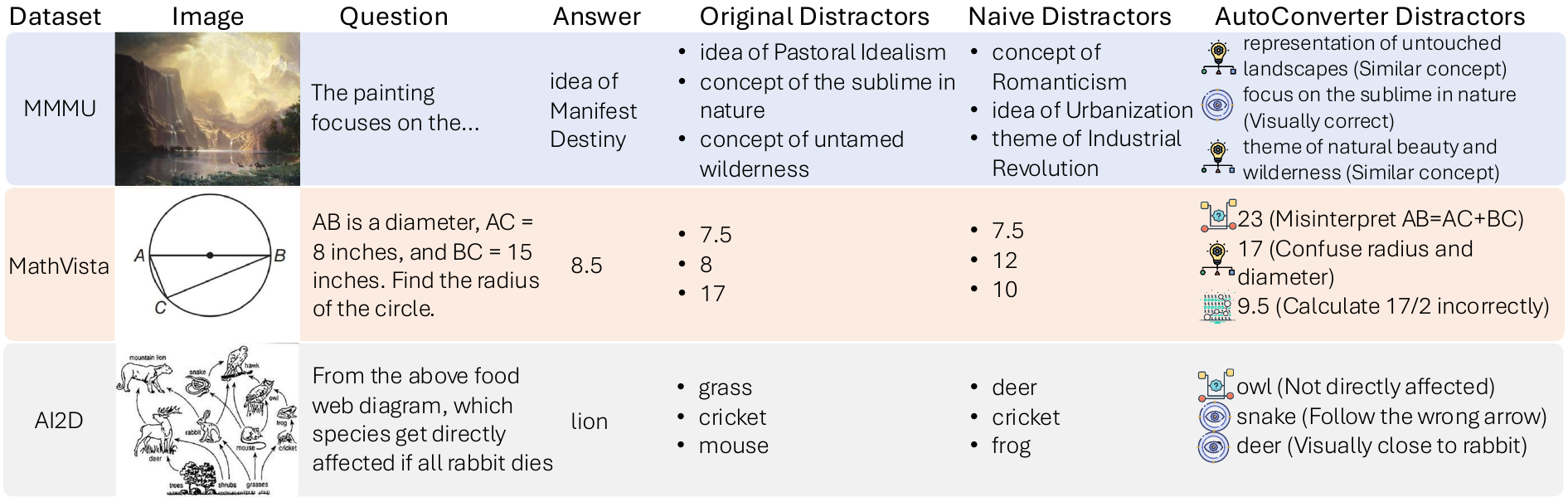}
    \vspace{-2em}
    \caption{\textbf{Qualitative comparison of the original questions, naive baseline-generated questions, and \textit{AutoConverter}-generated questions.} \textit{AutoConverter} simulates errors from different perspectives and produces correct and challenging multiple-choice questions. 
    }
    \label{fig:case_difficulty}
    \vspace{-1em}
\end{figure*}

\subsection{Increase Difficulty}

Difficulty is another critical factor in designing effective multiple-choice questions. Our goal is to prevent questions from being trivially answered due to obviously incorrect options or shortcuts, ensuring that each question retains the discriminative power needed to rigorously test VLM capabilities. Inspired by human strategies for creating challenging distractors, we introduce a multi-agent framework powered by GPT-4o. This framework iteratively generates and refines a large set of difficult distractors from multiple perspectives and ultimately selects those that pose the highest level of difficulty.

\textbf{Generating a diverse set of distractors.} We first create a large pool of candidate distractors based on common error types to capture a range of potential challenges. We categorize these error types as follows: concept misunderstanding, visual misinterpretation, reasoning error, data processing error, and question bias. Detailed definitions for each error type are provided in the Appendix. For each error type, GPT-4o proposes a set of distractors based on the image, question, and correct answer, along with accompanying rationales. This pool of candidates forms a comprehensive foundation for our selection process.

\textbf{Iterative refinement of distractors.} After generating initial distractors, we employ another GPT-4o agent as a reviewer to provide targeted feedback for each distractor. This reviewer assesses the plausibility and challenge level of each distractor given the context of the question and answer, offering suggestions for improvement. The specialized proposer agent then refines the distractors based on this feedback, ensuring that each distractor is as challenging as possible and removing those considered too close to the correct answer to maintain correctness.

\textbf{Selecting high-difficulty distractors.} From the refined pool of distractors with their creation rationales, we finally use GPT-4o as a selector to evaluate each candidate’s difficulty and correctness, selecting the most challenging distractors along with justifications for their selection. This process results in a final set of high-quality distractors that rigorously test VLM performance.

This process is illustrated in Figure~\ref{fig:method} (left). Our detailed ablation studies demonstrate that each of these agents contributes to improving the difficulty of the questions.

\begin{figure*}[!tb]
    \centering
    \includegraphics[width=\linewidth]{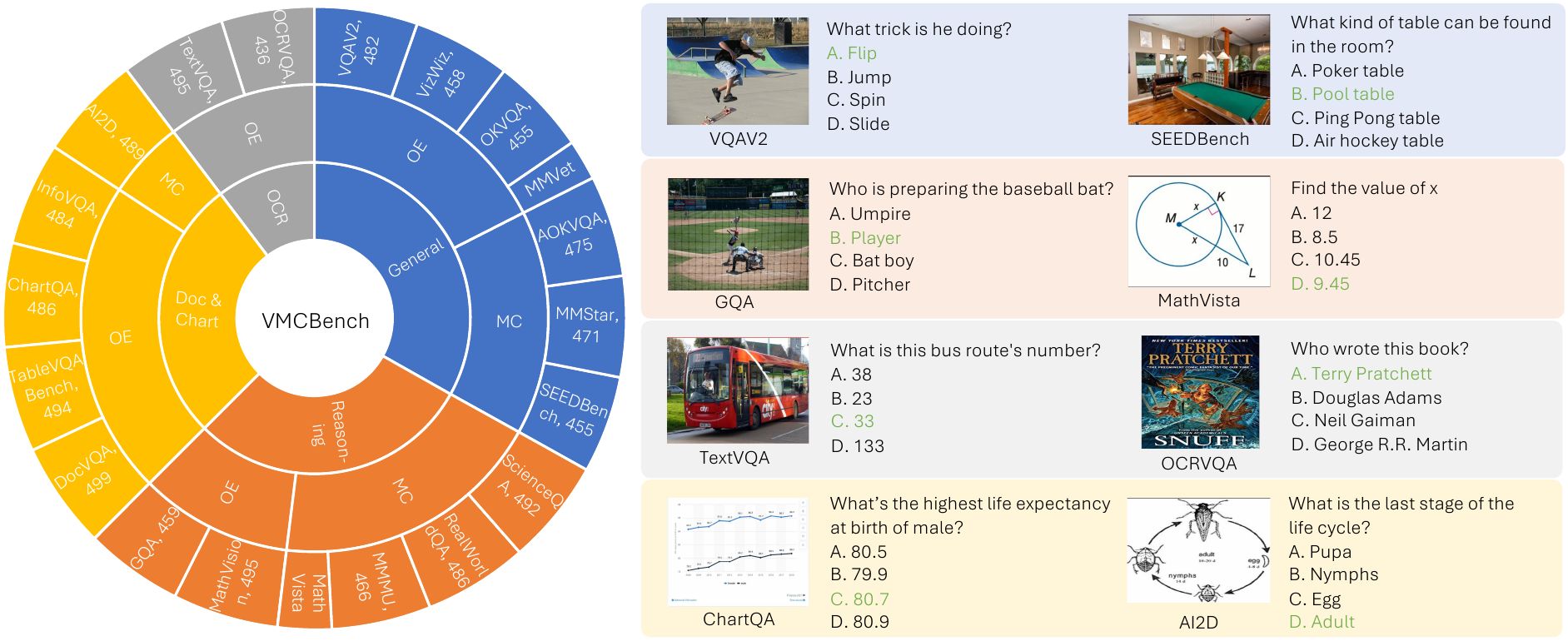}
    \vspace{-2em}
    \caption{
    \textbf{\textit{VMCBench} overview.} \textit{(Left)} \textit{VMCBench} is constructed by converting 12 open-ended (OE) and refining 8 multiple-choice (MC) VQA datasets into a unified multiple-choice format, with human validation ensuring correctness. The number of questions per dataset is listed. \textit{(Right)} Example questions from \textit{VMCBench}, showcasing diverse question types across multiple domains.
    }
    \vspace{-1em}

    \label{fig:case_benchmark}
\end{figure*}

\subsection{AutoConverter Results}

To evaluate the effectiveness of \textit{AutoConverter} in generating high-quality multiple-choice questions, we leverage three existing multiple-choice VQA datasets: MMMU~\cite{mmmu}, MathVista~\cite{mathvista}, and AI2D~\cite{ai2d}. We retain all questions with four choices and regenerate the distractors based on the image, question, and correct answer. We then compare \textit{AutoConverter}-generated distractors with the original human-crafted ones by evaluating various VLMs on both the original and converted datasets. 

\textbf{\textit{AutoConverter} generates correct multiple-choice questions.} 
Using a correctness evaluator and refiner pipeline, we ensure the accuracy of the generated questions. Our analysis shows that the generated questions maintain a correctness level comparable to that of the original human-crafted questions. After filtering to keep only questions with a correctness score of 5, large-scale human annotation indicates that only 3\% of questions are marked as incorrect for MMMU, MathVista, and AI2D. Among these, 52\% of errors are due to incorrect original answers, while only 48\% are introduced by the \textit{AutoConverter} process. These results demonstrate that most \textit{AutoConverter}-generated questions are accurate and suitable for reliable VLM evaluation.

\textbf{\textit{AutoConverter} generates challenging multiple-choice questions.} 
As shown in Figure~\ref{fig:difficulty}, \textit{AutoConverter} produces highly challenging questions, with a broad range of VLMs achieving similar or even lower accuracy on the generated distractors compared to the original human-crafted ones across MMMU, MathVista, and AI2D. Notably, MMMU is regarded as one of the most challenging VQA datasets, as it is sourced from exams and textbooks. These results highlight \textit{AutoConverter}’s capability not only in converting open-ended questions to multiple-choice but also in refining existing multiple-choice questions to enhance their difficulty. Furthermore, \textit{AutoConverter} could have applications beyond VLM evaluation, such as generating challenging questions for educational purposes.

\textbf{Each agent in \textit{AutoConverter} contributes to jointly improving correctness and difficulty.} To analyze each agent’s role in \textit{AutoConverter}, we conduct an ablation study by removing each component, as shown in Figure~\ref{tab:ablation}. First, removing specialized error proposers results in a decrease in difficulty, with an average increase of 1.6\% in relative accuracy. Next, omitting the reviewer for iterative distractor refinement leads to a 4.9\% relative increase in average VLM performance, indicating lower difficulty. Finally, removing the question evaluator and refiner, which serve as correctness safeguards, causes an 8.7\% relative decrease in correctness score. We also compare \textit{AutoConverter} with a naive baseline that uses the prompt “Please generate 3 distractors given the question, answer, and image.” As shown in Figure~\ref{tab:ablation}, \textit{AutoConverter} significantly outperforms this baseline in both correctness (2.2\% relative increase) and difficulty (11.5\% relative decrease in average VLM performance), demonstrating the effectiveness of our approach.

We provide qualitative comparisons of original and \textit{AutoConverter}-generated questions, as well as the naive baseline, in Figure~\ref{fig:case_difficulty}, which demonstrate the effectiveness of \textit{AutoConverter} in creating hard yet correct questions.

%% file: sections/4_benchmark.tex
\section{VMCBench: A Unified Multiple-Choice Visual Question Answering Benchmark}
\label{sec:benchmark}

Using our \textit{AutoConverter} method, detailed in \S\ref{sec:method}, we introduce \textit{VMCBench} --- a benchmark that unifies 20 existing visual question answering (VQA) datasets into a consistent multiple-choice format. \textit{VMCBench} spans a diverse array of visual and linguistic contexts, rigorously testing various model capabilities. By transforming open-ended questions into multiple-choice format, \textit{VMCBench} enhances vision language model evaluation by mitigating ambiguities while preserving the complexity of the tasks. This benchmark provides a reliable and reusable resource for the evaluation of future vision language models (VLMs).

\subsection{Benchmark Overview}

\textit{VMCBench} transforms 20 widely-used VQA datasets into a unified multiple-choice benchmark. These datasets can be broadly categorized to assess general capabilities of VLMs (VQAv2~\cite{vqav2}, OKVQA~\cite{okvqa}, MMVet~\cite{mmvet}, VizWiz~\cite{vizwiz}, A-OKVQA~\cite{aokvqa}, MMStar~\cite{mmstar}, SEEDBench~\cite{seedbench}), reasoning capabilities (MathVision~\cite{mathvision}, GQA~\cite{gqa}, MMMU~\cite{mmmu}, RealWorldQA~\cite{realworldqa}, MathVista~\cite{mathvista}, ScienceQA~\cite{scienceqa}), OCR tasks (OCRVQA~\cite{ocrvqa}, TextVQA~\cite{textvqa}), and document and chart understanding (DocVQA~\cite{docvqa}, InfoVQA~\cite{infovqa}, ChartQA~\cite{chartqa}, TableVQABench~\cite{tablevqa}, AI2D~\cite{ai2d}).

Among these, 12 datasets consist of open-ended questions, which we convert into the multiple-choice format to facilitate evaluation. The remaining 8 datasets are already in multiple-choice format; for these, we apply \textit{AutoConverter} to refine distractors, increasing their difficulty. These refined datasets can further test model robustness and help identify dataset contamination~\cite{recht2019imagenet,hendrycks2021many,zhang2024careful,orenproving}.

For each dataset, we randomly sampled up to 500 questions, as recent studies suggest this sample size may suffice for evaluating model performance~\cite{polo2024tinybenchmarks,perlitz2024efficient}, resulting in a total of 9,450 questions across the 20 datasets. We apply \textit{AutoConverter} to these 9,450 questions.

To ensure the correctness of the converted questions, \textit{AutoConverter} incorporates an internal correctness evaluator that assigns a correctness score to each converted question, ensuring that only one correct answer exists. Among the 9,450 converted questions, 103, 162, 399, 635, and 8,151 questions received correctness scores of 1--5, respectively. Six PhD student annotators reviewed all questions with correctness scores below 5 and randomly sampled 1,101 questions with a score of 5, resulting in a total of 2,400 annotated questions. The correctness rates for questions with scores of 1--5 were 51\%, 51\%, 63\%, 84\%, and 95\%, respectively. This increasing trend in the correctness rate demonstrates the reliability of the correctness evaluation.

To gain further insight, we annotated the source of errors, distinguishing between those stemming from ground-truth answer inaccuracies (errors in the original dataset) and those introduced by \textit{AutoConverter}. Our analysis revealed that 59\% of errors were due to issues in the ground-truth answers rather than \textit{AutoConverter}. Notably, if all original open-ended questions were error-free, \textit{AutoConverter} would have introduced only 2\% errors among questions with a correctness score of 5, highlighting the reliability of our approach. Therefore, we recommend retaining only questions with a correctness score of 5 for future benchmark construction. We removed all questions identified as incorrect, resulting in a final total of 9,018 questions.

\input{tables/vmcbench_short}

\textit{VMCBench} spans a diverse range of visual and linguistic contexts and includes a broad array of question types, offering a comprehensive framework to evaluate VLMs across multiple vision language tasks. It provides a unified interface for efficient and accurate assessment of VLM capabilities. We split the \textit{VMCBench} into 1,000 validation questions with provided answers and 8,018 test questions without answers. An online test server is available to support future VLM development. The question distribution and examples of converted questions are shown in Figure~\ref{fig:case_benchmark}.

\subsection{Evaluation Results}

We evaluated 33 state-of-the-art VLMs on the \textit{VMCBench}, including GPT-4~\cite{gpt4}, Gemini-1.5~\cite{team2023gemini}, Claude-3.5~\cite{claude}, Qwen2-VL~\cite{qwen2vl}, Molmo~\cite{deitke2024molmo}, Cambrian~\cite{cambrian}, VILA~\cite{vila}, CogVLM~\cite{cogvlm2}, Phi~\cite{phi}, Idefics2~\cite{idefics2}, DeepSeek-VL~\cite{deepseekvl}, PaliGemma~\cite{paligemma}, LLaVA1.5~\cite{llava1.5}, Chameleon~\cite{chameleon}, and InstructBLIP~\cite{instructblip}. The evaluation is conducted in the zero-shot setting with the prompt ``Question: \{question\} Options: A. \{A\} B. \{B\} C. \{C\} D. \{D\} Answer with the option's letter from the given choices directly.'' The results are summarized in Table~\ref{tab:vlm_results}. Detailed results are in Appendix.

Our findings reveal several key insights:

\textbf{The gap between private and public models is narrowing.} As shown in Table~\ref{tab:vlm_results}, the best-performing model on \textit{VMCBench} is the publicly available Qwen2-VL-72B, which achieves 85.0\% overall accuracy---surpassing the best private model, GPT-4o, which scores 80.3\%.

\textbf{VLM development is advancing rapidly.} Another notable trend is the substantial performance gain from InstructBLIP-7B to Qwen2-VL-72B, with accuracy nearly doubling from 42.1\% to 85.0\%. These models were released in 2023 and 2024, respectively, highlighting the rapid pace of progress in VLMs.

\textbf{Scaling up models improves performance.} As expected, scaling up model sizes leads to performance gains. For instance, across different model families (such as Qwen, Molmo, VILA, Cambrian, DeepSeek, and LLaVA), we observe a clear improvement with larger models, similar to scaling trends in language models~\cite{kaplan2020scaling, brown2020language}.

\textbf{Diverse datasets reveal model limitations.} While Phi-3.5 outperforms Phi-3 on reasoning tasks like MMMU and MathVista, it underperforms on OCR and doc/chart tasks, lowering its overall accuracy. This highlights the need for broad evaluation. \textit{VMCBench} enables this by unifying diverse datasets into a standardized multiple-choice format.

We believe that \textit{VMCBench} is a valuable benchmark that will play a significant role in advancing VLM development by facilitating simple, fast, and accurate model evaluation. 

%% file: tables/vmcbench_short.tex
\begin{table}[!tb]
\centering
\rowcolors{2}{white}{light-light-gray}
\setlength\tabcolsep{1pt}
\renewcommand{\arraystretch}{1.1}
\small
\begin{tabular}{lccccc}
\toprule
\textbf{Model} & \textbf{General} & \textbf{Reasoning} & \textbf{OCR} & \textbf{Doc\&Chart} & \textbf{Avg.} \\
\midrule
Qwen2-VL-72B & 88.5 & 72.6 & 96.8 & 90.1 & 85.0 \\
GPT-4o & 85.2 & 66.9 & 96.4 & 83.1 & 80.3 \\
Molmo-72B & 82.9 & 66.6 & 94.7 & 81.1 & 78.7 \\
Qwen2-VL-7B & 84.5 & 62.7 & 96.4 & 80.1 & 78.1 \\
Claude-3.5-Sonnet & 81.3 & 62.8 & 93.4 & 84.6 & 77.8 \\
Cambrian-34B & 83.7 & 65.9 & 95.7 & 73.3 & 77.0 \\
Gemini-1.5-Pro & 79.6 & 64.7 & 92.6 & 72.6 & 74.7 \\
VILA1.5-40B & 82.5 & 65.3 & 93.2 & 67.4 & 74.7 \\
GPT-4o-Mini & 80.9 & 58.8 & 93.8 & 74.8 & 74.0 \\
Qwen2-VL-2B & 77.9 & 55.8 & 93.1 & 72.5 & 71.5 \\
CogVLM2-19B & 78.1 & 55.6 & 92.3 & 72.6 & 71.4 \\
Phi-3-Vision & 74.1 & 56.4 & 90.6 & 73.8 & 70.3 \\
Cambrian-13B & 79.3 & 54.6 & 92.3 & 66.6 & 70.0 \\
Cambrian-8B & 77.9 & 56.4 & 91.0 & 65.4 & 69.6 \\
Molmo-7B-D & 73.2 & 55.5 & 91.7 & 72.1 & 69.5 \\
Idefics2-8B & 77.8 & 55.8 & 92.7 & 61.8 & 68.7 \\
Molmo-7B-O & 72.6 & 54.3 & 88.5 & 68.9 & 67.8 \\
Phi-3.5-Vision & 71.4 & 55.3 & 87.2 & 68.6 & 67.4 \\
VILA1.5-13B & 74.6 & 54.1 & 85.3 & 50.2 & 63.4 \\
DeepSeek-VL-7B & 73.2 & 52.6 & 85.8 & 52.9 & 63.2 \\
Molmo-1B & 69.4 & 50.1 & 87.4 & 60.2 & 63.1 \\
CogVLM-17B & 72.3 & 48.8 & 77.8 & 54.6 & 61.3 \\
VILA1.5-8B & 72.4 & 51.8 & 81.8 & 46.5 & 60.7 \\
Gemini-1.5-Flash & 59.7 & 53.8 & 79.9 & 56.3 & 59.1 \\
PaliGemma-3B & 71.7 & 51.3 & 53.1 & 53.0 & 59.0 \\
VILA1.5-3B & 70.3 & 48.0 & 78.9 & 42.3 & 57.5 \\
DeepSeek-VL-1.3B & 68.9 & 43.6 & 79.5 & 43.8 & 56.1 \\
LLaVA1.5-13B & 66.4 & 46.2 & 75.8 & 37.0 & 53.9 \\
LLaVA1.5-7B & 63.6 & 44.7 & 74.0 & 35.0 & 51.8 \\
Chameleon-30B & 53.1 & 41.2 & 48.0 & 33.9 & 44.2 \\
InstructBLIP-7B & 55.1 & 35.2 & 47.7 & 29.9 & 42.1 \\
InstructBLIP-13B & 54.8 & 34.7 & 48.7 & 26.5 & 41.1 \\
Chameleon-7B & 41.0 & 34.7 & 42.3 & 29.6 & 36.4 \\
\bottomrule
\end{tabular}
\vspace{-1em}
\caption{\textbf{Performance of 33 VLMs on \textit{VMCBench} test set.}}
\vspace{-2em}
\label{tab:vlm_results}
\end{table}

%% file: sections/6_conclusion.tex
\section{Conclusion}

We address the limitations of open-ended visual question answering benchmarks for evaluating vision language models (VLMs) by introducing \textit{AutoConverter}, a multi-agent system that transforms questions into multiple-choice format with challenging distractors. \textit{AutoConverter} generates questions that rival or surpass human-written ones in difficulty. Using it, we construct \textit{VMCBench}, a benchmark of 9,018 multiple-choice questions that sets a new standard for consistent and rigorous VLM evaluation.

%% file: appendix_arxiv.tex
\section*{Acknowledgements} 

We thank Ross Girshick for discussing the idea behind this project. We also thank OpenAI's Researcher Access Program (ID 0000005368) for granting us API access. Serena Yeung-Levy is a Chan Zuckerberg Biohub --- San Francisco Investigator.

\section*{Broader Impacts and Ethics Statement}
\label{sec:impact}

\textit{AutoConverter} and \textit{VMCBench} simplify and standardize vision language model (VLM) evaluation, providing a valuable tool for advancing VLM development through scalable, consistent, and reproducible evaluation. Beyond VLM evaluation, \textit{AutoConverter} can be applied to other domains, such as education, to generate challenging and high-quality questions. However, human verification is essential to ensure the generated questions align with their intended purpose, maintain proper value, and achieve the appropriate level of difficulty without introducing biases.

\section*{Reproducibility Statement}
\label{sec:reproduce}

We provide an open-source implementation of our work at \url{https://github.com/yuhui-zh15/AutoConverter}. This will enable researchers to reproduce all the experiments in paper and conduct their own analyses.

\section*{Limitations}
\label{sec:limitation}

\textit{AutoConverter} achieves high question correctness, but 5\% of the questions with the highest correctness scores still contain errors, although half of which are due to flaws in the original datasets. Moreover, while \textit{VMCBench} includes a diverse range of models and datasets, its coverage remains incomplete. Moving forward, we aim to leverage \textit{AutoConverter} to expand dataset and model coverage, addressing evolving needs in VLM evaluation.

\section*{Summary of Appendix}

\begin{itemize}
\item In \S\ref{sec:supp3}, we present further analysis of the evaluation failures for open-ended questions discussed in \S3.
\item In \S\ref{sec:supp4}, we provide supplementary details on the \textit{AutoConverter} framework described in \S4.
\item In \S\ref{sec:supp5}, we elaborate more details of \textit{VMCBench} as outlined in \S5.
\end{itemize}

\section{Supplementary Section 3}
\label{sec:supp3}

\subsection{Examples of Rule-based Evaluation Failures}

In the main paper, we demonstrated that rule-based evaluation of open-ended questions leads to poor evaluation outcomes. Table~\ref{tab:failures-rule} provides six examples of rule-based evaluation failures, clearly illustrating that these methods fail to account for semantic similarity and penalize formatting errors, resulting in highly inaccurate evaluation results.

\subsection{Examples of Model-based Evaluation Failures}

In the main paper, we demonstrated that model-based evaluation is sensitive to model versions. To investigate this further, Table~\ref{tab:failures-model} presents six examples of inconsistencies caused by different model-based evaluations. We observe that GPT-4o-0806 often assigns a perfect score of 1.0 for similar predictions and answers, whereas GPT-4o-0513 tends to assign a score of 0.9. This behavior variation introduces significant differences in evaluation results and raises concerns about reproducibility in future research.

\section{Supplementary Section 4}
\label{sec:supp4}

\subsection{Prompts for \textit{AutoConverter}}

In the main paper, we introduced the agentic system of \textit{AutoConverter}, comprising the proposer, reviewer, selector, evaluator, and refiner agents. This system creates a large pool of high-quality distractor options to increase question difficulty while ensuring correctness in the converted multiple-choice questions.

Detailed prompts for the proposers, designed to create distractors addressing \textit{vision}, \textit{reasoning}, \textit{data}, \textit{concept}, and \textit{bias} errors, are shown in Figures~\ref{fig:prompt-visual}, \ref{fig:prompt-reasoning}, \ref{fig:prompt-data}, \ref{fig:prompt-concept}, and \ref{fig:prompt-bias}, respectively. These prompts are carefully crafted to align with common types of human errors, as defined in the corresponding sections.

Figure~\ref{fig:prompt-reviewer} presents prompts for the \textit{reviewer}, whose feedback iteratively refines the distractors to improve their quality. The \textit{selector} prompts in Figure~\ref{fig:prompt-selector} guide the selection of the three most challenging distractors to enhance question difficulty.

Figures~\ref{fig:prompt-evaluator} and \ref{fig:prompt-refiner} show prompts for the \textit{evaluator} and \textit{refiner}, respectively. This iterative process ensures the correctness of the generated questions, guaranteeing that there is only one correct answer.

\subsection{\textit{AutoConverter} Results for Additional Datasets}

\begin{figure*}[!tb]
    \includegraphics[width=\linewidth]{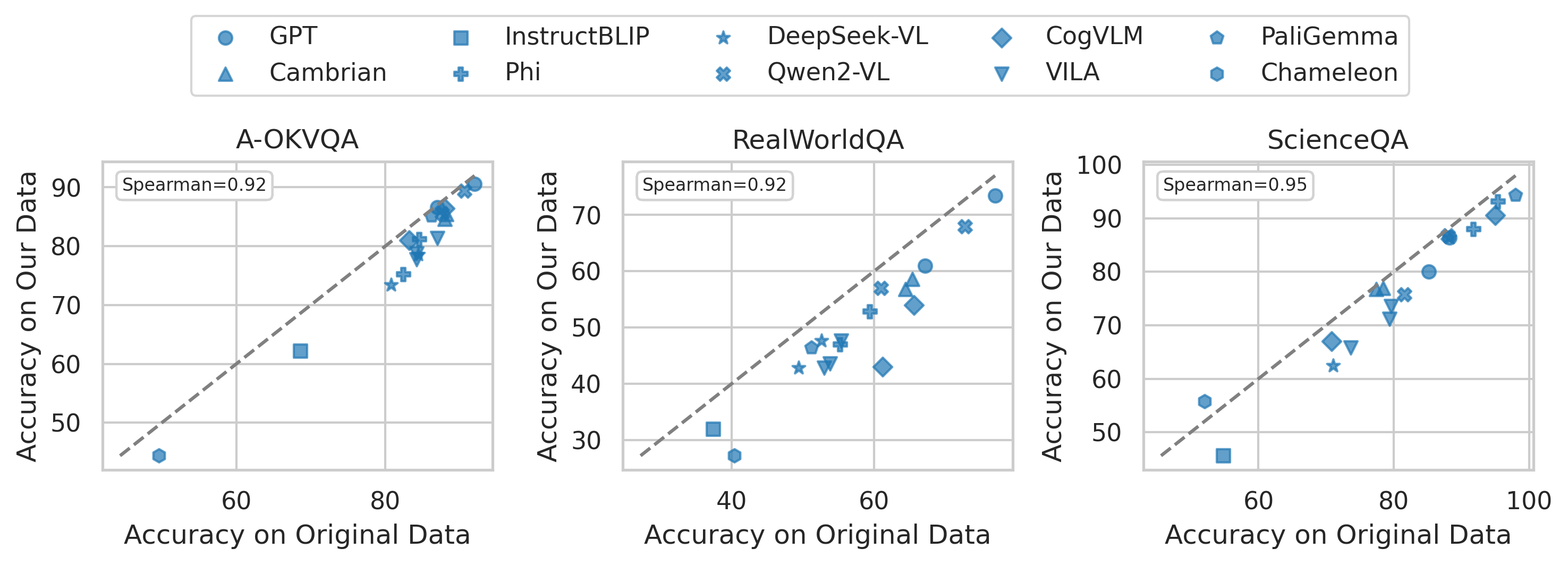}
    \includegraphics[width=0.9\linewidth]{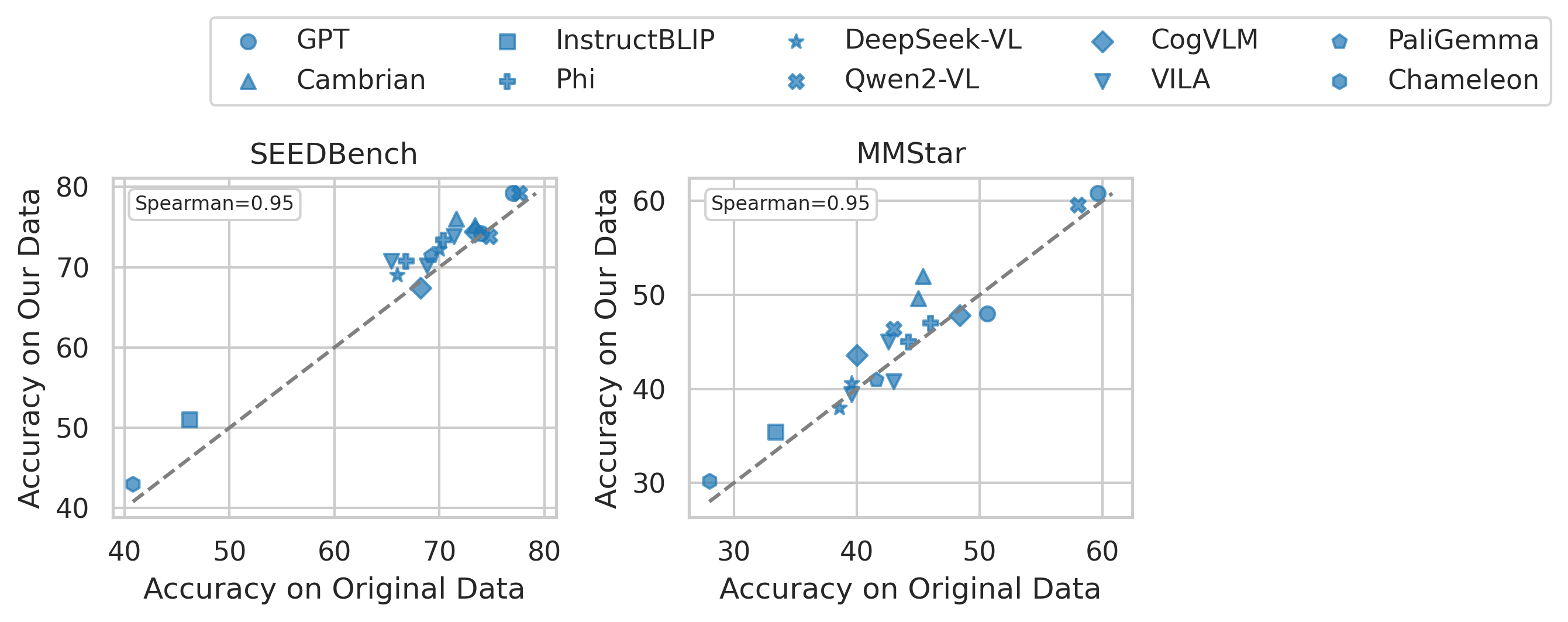}
    \caption{\textbf{\textit{AutoConverter} generates challenging multiple-choice questions.} 
    Using \textit{AutoConverter}, we generated distractors for questions and answers from five multiple-choice datasets: A-OKVQA, RealWorldQA, ScienceQA, SEEDBench, and MMStar, and compared them with original human-created distractors. We evaluated various VLMs on both the \textit{AutoConverter}-generated and the original questions, finding that VLMs consistently achieved similar or even lower accuracy on the \textit{AutoConverter}-generated questions compared to the original ones.
    }
    \vspace{-1em}
    \label{fig:superhuman}
\end{figure*}

In the main paper, we demonstrated \textit{AutoConverter}'s ability to generate challenging multiple-choice questions for three datasets: MMMU, MathVista, and AI2D. Figure~\ref{fig:superhuman} extends this analysis to five additional datasets with human-created distractors: A-OKVQA, RealWorldQA, ScienceQA, SEEDBench, and MMStar. Across these datasets, VLMs consistently achieved similar or lower accuracy on \textit{AutoConverter}-generated questions compared to the original ones, demonstrating the system's capability to produce highly challenging multiple-choice questions.

\subsection{\textit{AutoConverter} Results for Different Models}

In the main paper, we constructed \textit{AutoConverter} using GPT-4o. A potential concern is whether this choice introduces bias into the generated questions.

To examine this, we used three state-of-the-art proprietary VLMs—GPT-4o, Claude-3.5-Sonnet, and Gemini-1.5-Pro—to generate questions. We evaluated various VLMs on these questions and computed the correlations of their performance rankings. If no bias exists, we expect high correlations across all question sets, as they should reflect the true discriminative power of the models.

\begin{figure*}[!tb]
    \includegraphics[width=\linewidth]{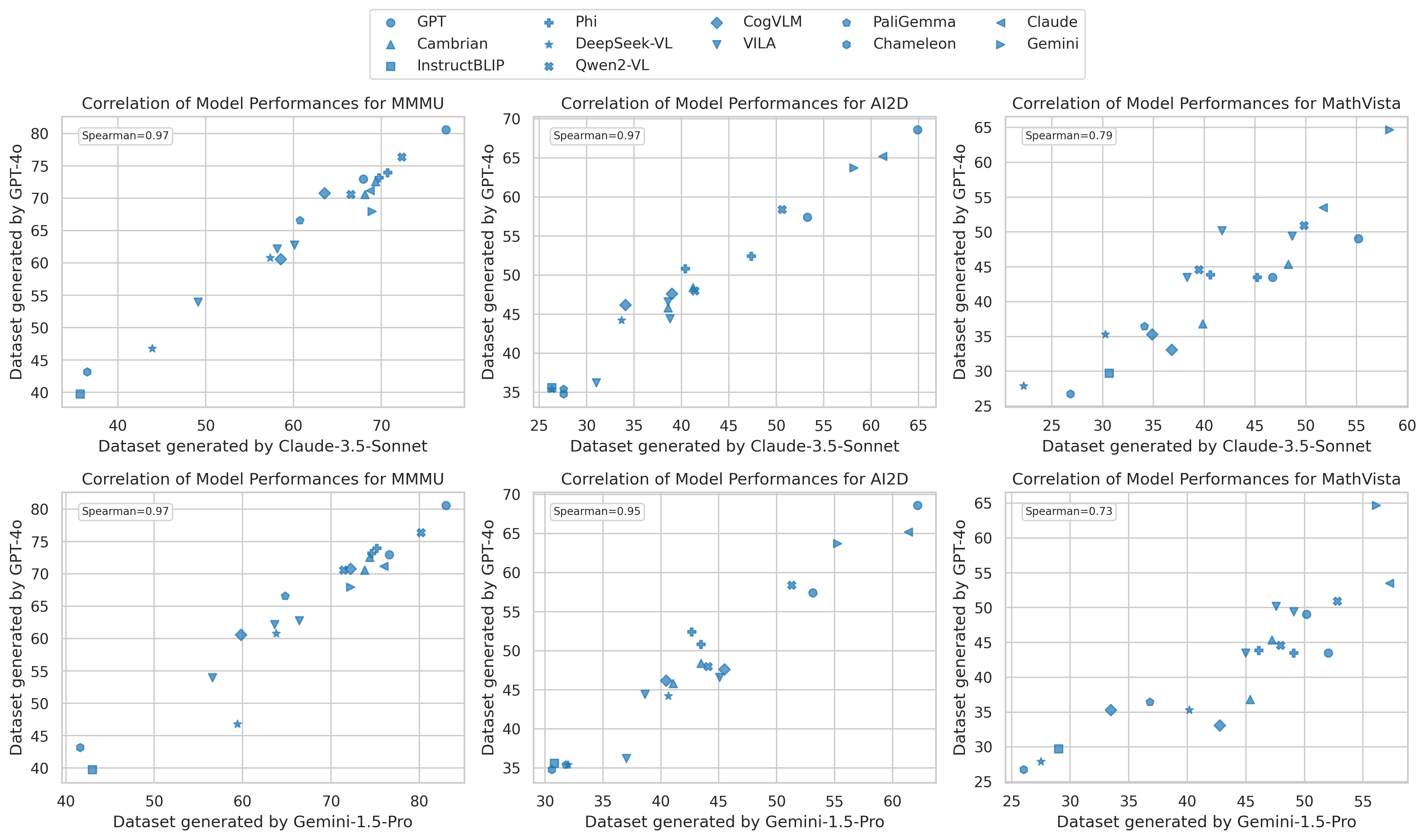}
    \caption{\textbf{\textit{AutoConverter} results for different models.} To examine whether GPT-4o used in \textit{AutoConverter} introduces model bias, we used three state-of-the-art proprietary VLMs—GPT-4o, Claude-3.5-Sonnet, and Gemini-1.5-Pro—to generate questions. We evaluated 18 VLMs on these questions and computed the correlations of their performance rankings. We observe high correlations across all question sets, indicating no model bias exists, and these questions reflect the true discriminative power of the models.}
    \vspace{-1em}
    \label{fig:other_models}
\end{figure*}

Figure~\ref{fig:other_models} supports our hypothesis. The rankings derived from GPT-4o, Claude-3.5-Sonnet, and Gemini-1.5-Pro questions exhibit near-perfect correlations (0.90 Spearman correlation averaged over three datasets), indicating that the choice of generator does not introduce significant bias.

\subsection{Correlation between Open-ended and Multiple-choice Questions}

In the main paper, we highlighted the challenges of accurately evaluating open-ended questions. Therefore, we propose an alternative solution to convert open-ended questions into a multiple-choice format. A key question here is whether converting open-ended questions into multiple-choice format preserves their discriminative power.

To address this, we treat model-based evaluation of open-ended questions as a proxy for ground-truth evaluation. Specifically, if a model-based evaluator determines that model A outperforms model B, we consider this ranking correct. This assumption is valid because model-based evaluations have a high correlation with human judgments, albeit with instability across versions. By using the same GPT version as the evaluator, we eliminate such instability.

\begin{figure*}[!tb]
    \includegraphics[width=\linewidth]{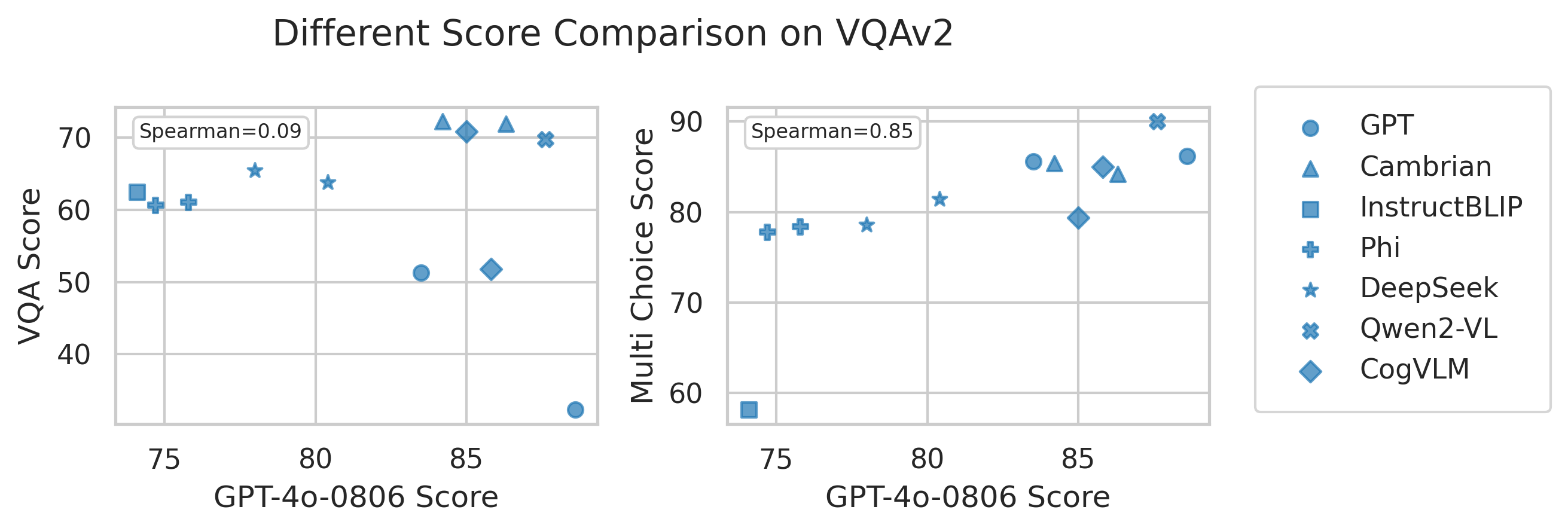}
    \includegraphics[width=\linewidth]{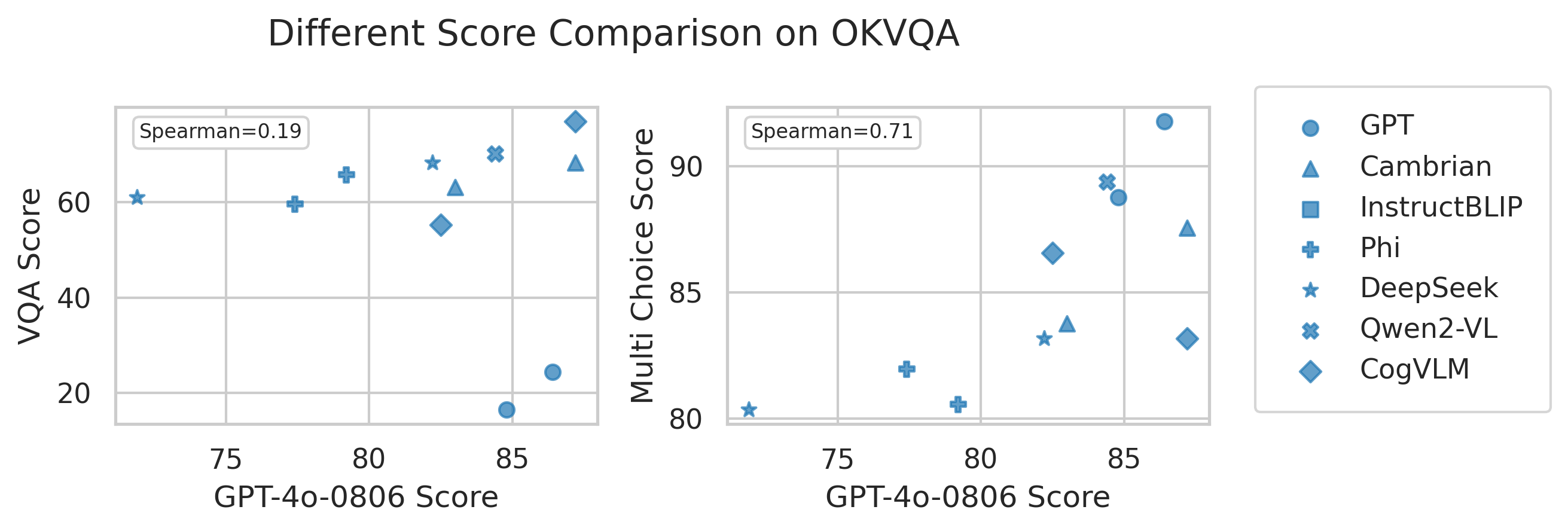}
    \includegraphics[width=\linewidth]{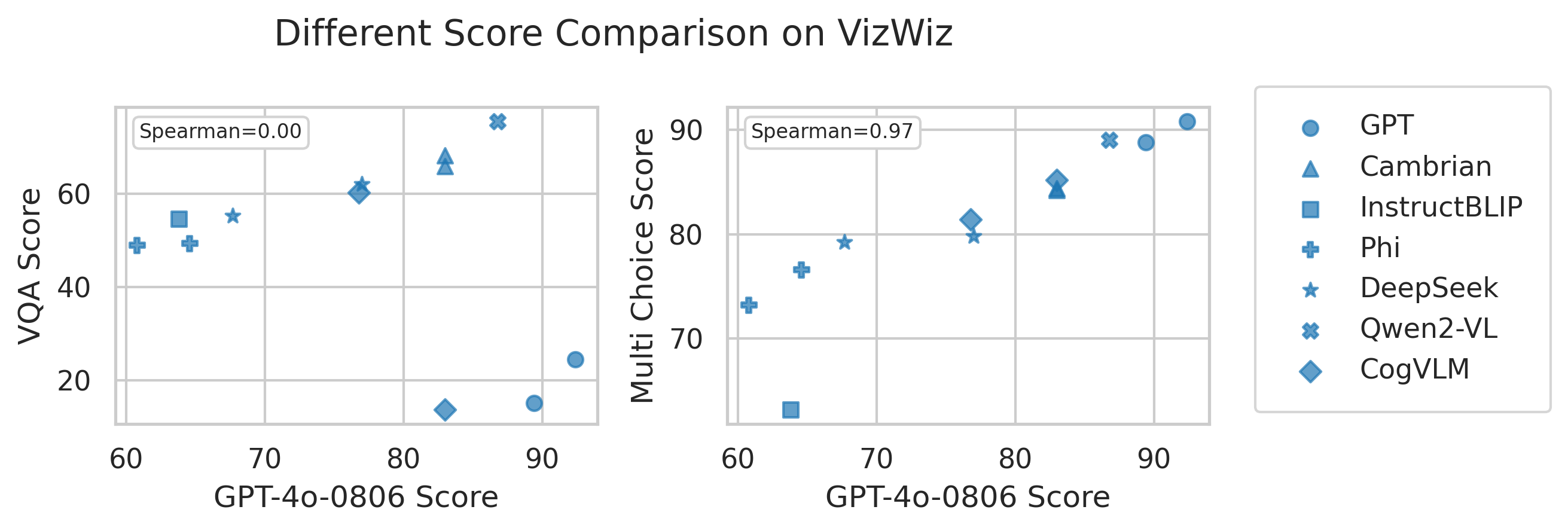}
    \caption{\textbf{Converting to multiple-choice questions improves evaluation accuracy and retains discriminative power.} We treat model-based evaluation of open-ended questions as a proxy for ground-truth evaluation. We compare the correlation between model-based evaluation of open-ended questions and rule-based evaluation of multiple-choice questions against the correlation between model-based and rule-based evaluation of open-ended questions. We find that the correlation for multiple-choice questions is significantly higher compared to rule-based open-ended evaluations, demonstrating that converting open-ended questions into multiple-choice format preserves their discriminative power and simplifies the evaluation.}
    \vspace{-1em}
    \label{fig:corr}
\end{figure*}

We compare the correlation between model-based evaluation of open-ended questions and rule-based evaluation of multiple-choice questions against the correlation between model-based and rule-based evaluation of open-ended questions. Our findings, shown in Figure~\ref{fig:corr}, reveal that the correlation for multiple-choice questions is significantly higher—0.85, 0.71, and 0.97 for VQAv2, OKVQA, and VizWiz, respectively—compared to rule-based open-ended evaluations, which achieve correlations of 0.09, 0.19, and 0.00. This demonstrates that converting to multiple-choice questions improves evaluation accuracy and retains discriminative power.

\section{Supplementary Section 5}
\label{sec:supp5}

\subsection{Human Evaluation Results on \textit{VMCBench}}

\textit{VMCBench} provides a scalable, consistent, and reproducible benchmark for evaluating and advancing VLMs. The current best-performing model, GPT-4o, achieves an accuracy of 80.6\%. 

To assess the remaining room for improvement, we conducted a human evaluation of \textit{VMCBench}. Human experts achieved an accuracy of 91.7\%, highlighting significant opportunities for further model improvement. Among the 8.3\% human errors, approximately three-quarters of the questions require extensive knowledge, while a quarter are ambiguous and unanswerable.

\subsection{Full Evaluation Results on \textit{VMCBench}}

In the main paper, we reported model performances grouped by benchmarked capabilities. Table~\ref{tab:vlm_results_full} presents the full evaluation results for 33 VLMs across 20 individual benchmarks, further validating our conclusions.

\subsection{Scaling Trends on \textit{VMCBench}}

\begin{figure*}[!tb]
    \centering
    \includegraphics[width=0.6\linewidth]{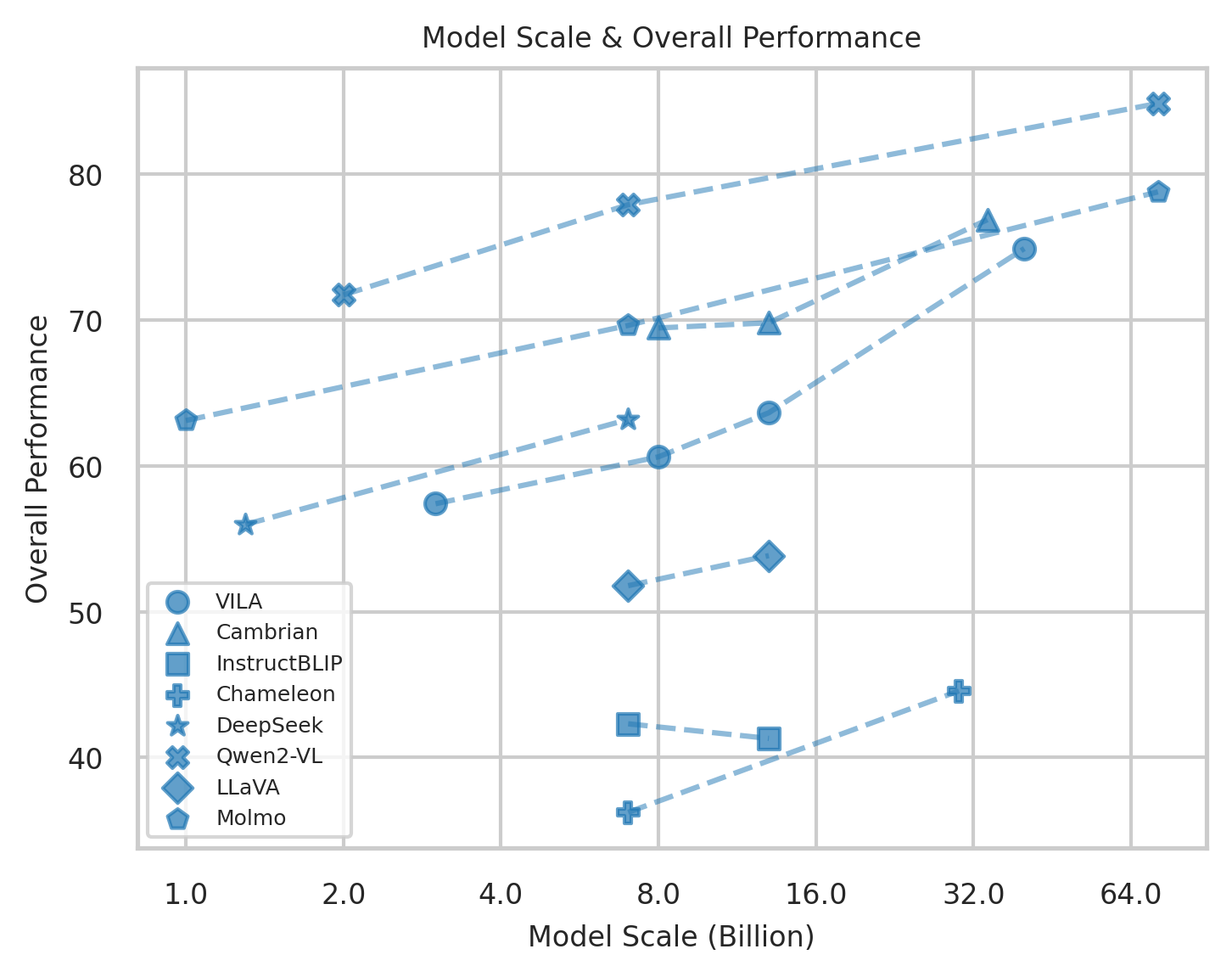}
    \includegraphics[width=0.4\linewidth]{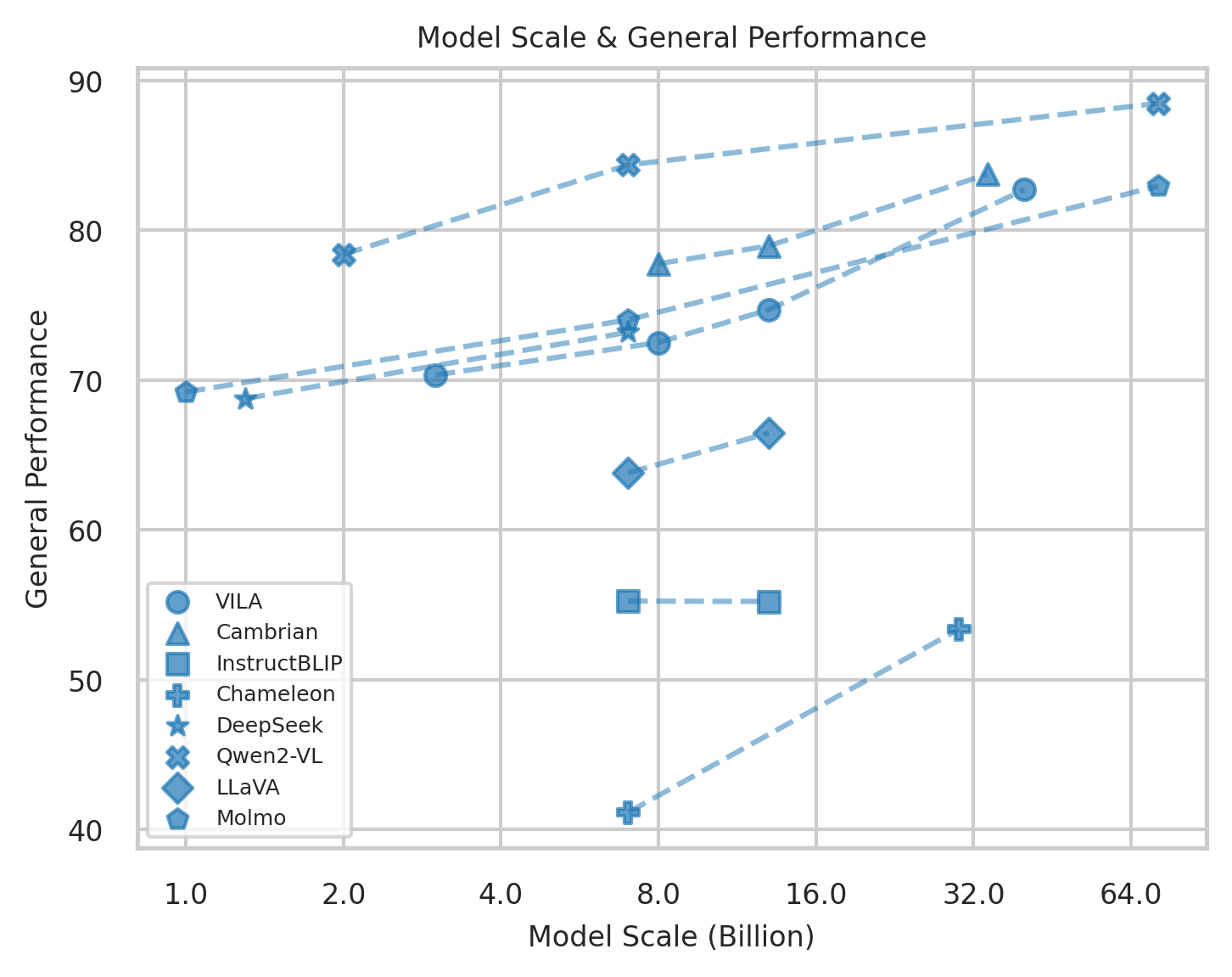}
    \includegraphics[width=0.4\linewidth]{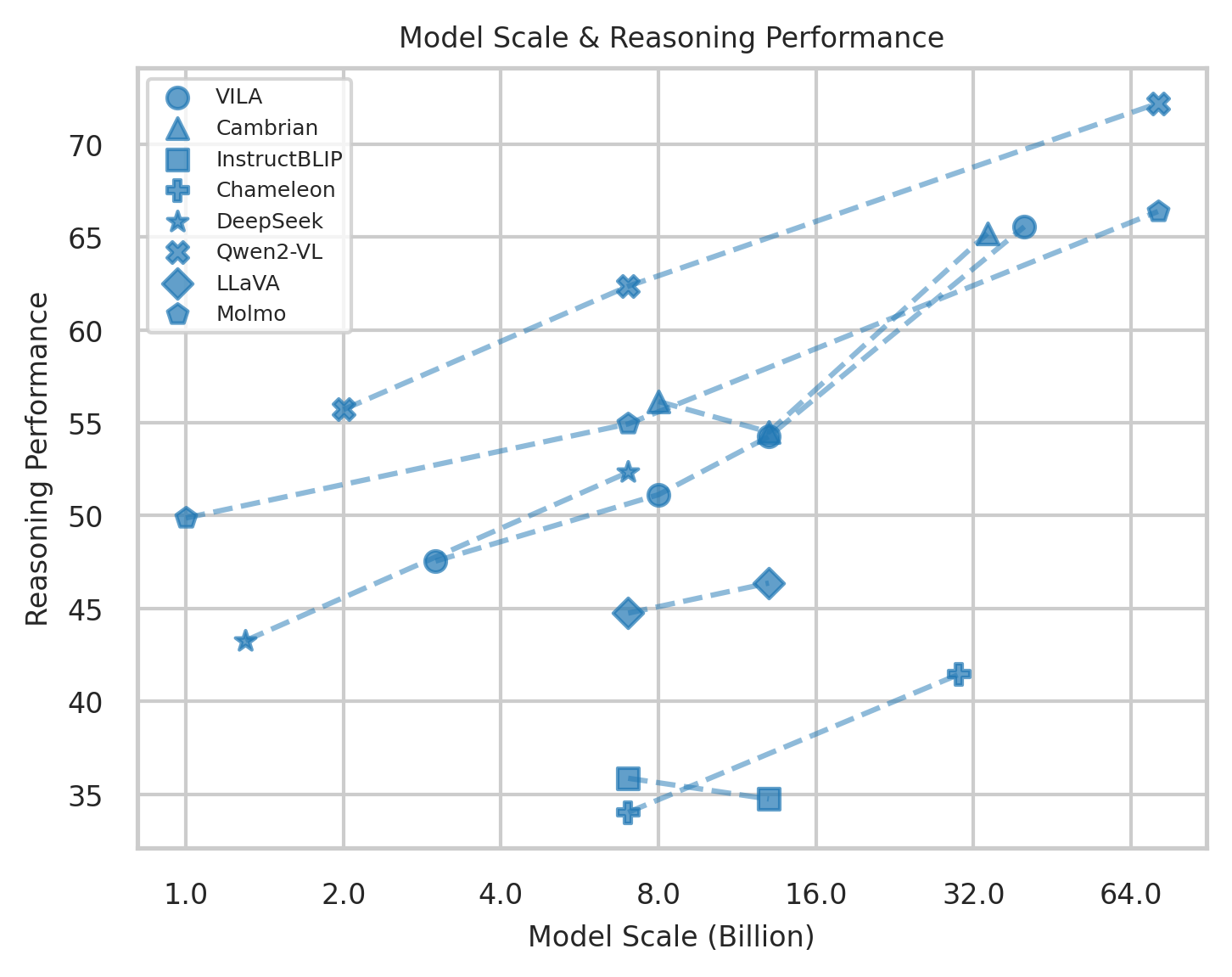}
    \includegraphics[width=0.4\linewidth]{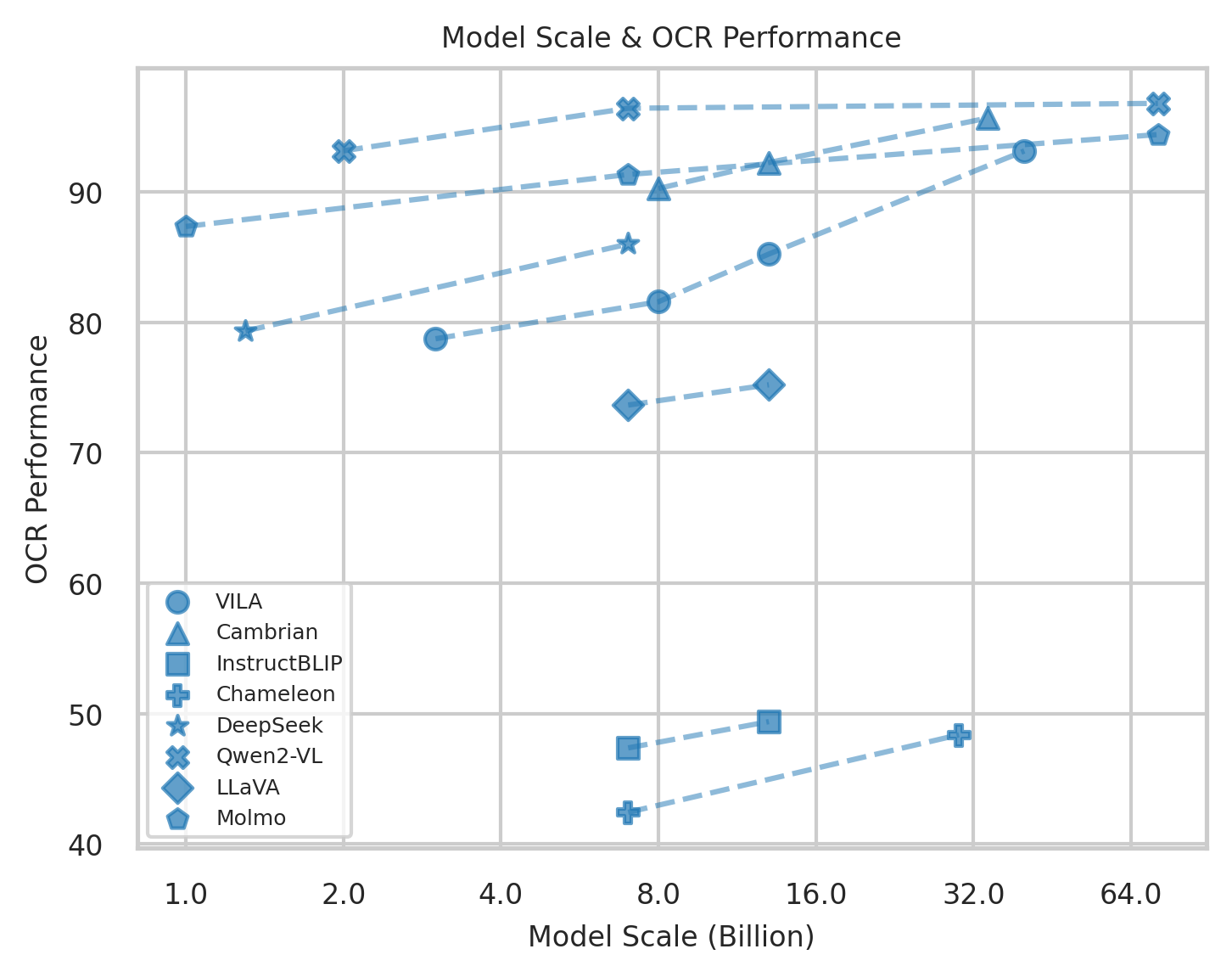}
    \includegraphics[width=0.4\linewidth]{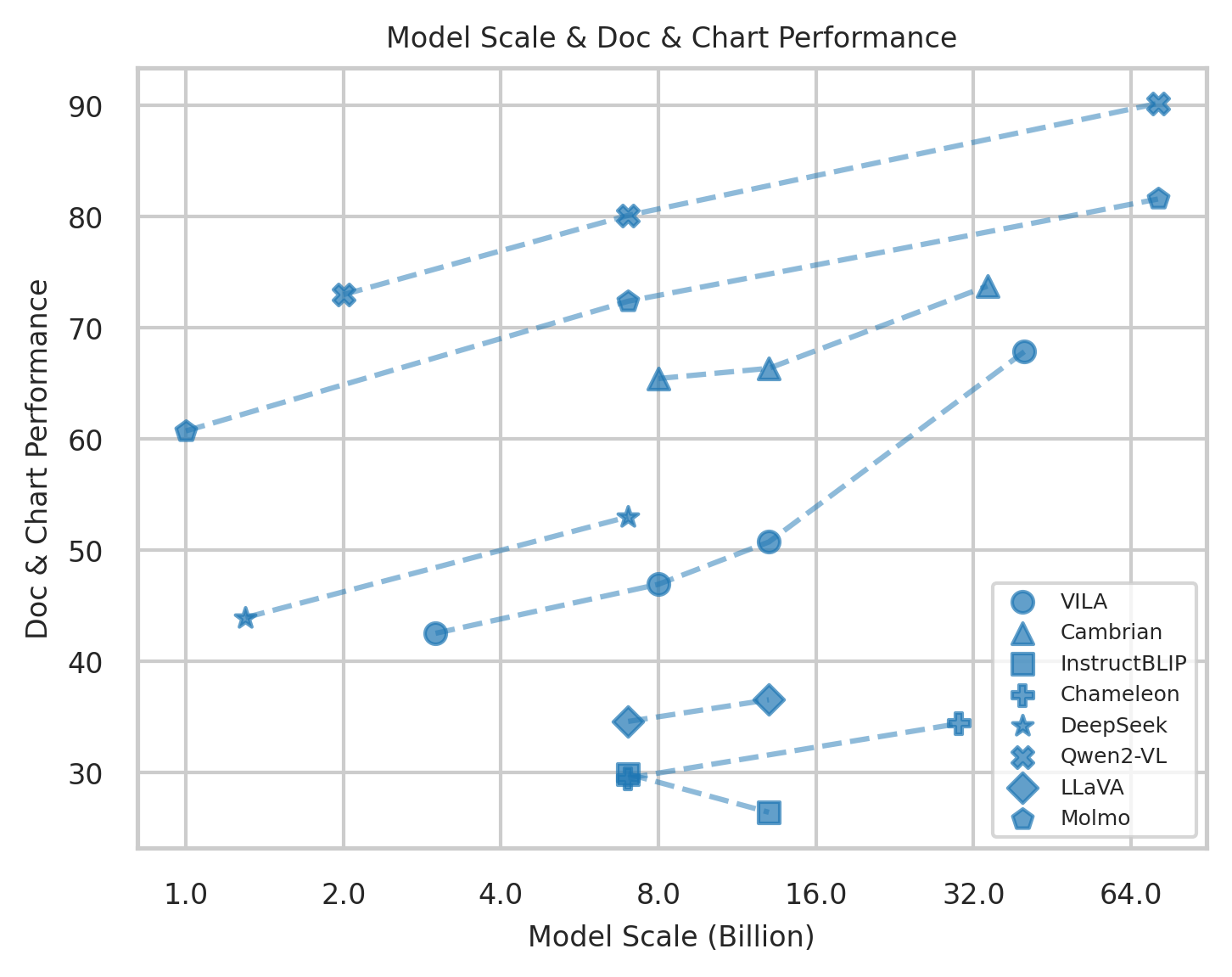}
    \caption{\textbf{Scaling trends on \textit{VMCBench}.}  We observe a clear \textbf{log-linear} scaling trend across most VLM families, indicating that \textit{VMCBench} offers a smooth evaluation gradient for varying capabilities. }
    \vspace{-1em}
    \label{fig:scaling}
\end{figure*}

Scaling laws are a cornerstone of model development, demonstrating that larger models typically achieve better performance. To explore this, we plotted the scaling trends of VLM families in log scale based on known model sizes, as shown in Figure~\ref{fig:scaling}.

Surprisingly, we observe a clear \textbf{log-linear} scaling trend across most VLM families, indicating that \textit{VMCBench} offers a smooth evaluation gradient for varying capabilities. Certain model families outperform others, leaving further exploration of these scaling trends to future work.

\subsection{Option Permutation Sensitivity on \textit{VMCBench}}

To assess whether VLMs are sensitive to option permutation, we randomly shuffled all correct options twice using two different random seeds. We observed no significant difference in performance on \textit{VMCBench} (Table~\ref{tab:shuffle2}). This result demonstrates the robustness of multiple-choice evaluation in \textit{VMCBench}.

\begin{table}[htbp]
    \centering
    \small
    \rowcolors{2}{white}{light-light-gray}
    \begin{tabular}{lcccccc}
    \toprule
       Question & $M_1$ & $M_2$ & $M_3$ & $M_4$ & $M_5$ & $M_6$ \\
    \midrule
        Original & 59.7 & 52.6 & 68.3 & 78.8 & 78.1 & 80.8 \\
        Shuffled 1 & 59.8 & 52.7 & 68.0 & 78.6 & 77.7 & 81.0 \\
        Shuffled 2 & 59.9 & 52.5 & 67.9 & 78.7 & 78.2 & 80.3  \\
    \bottomrule
    \end{tabular}
    \vspace{-1em}
    \caption{\textbf{Performance on \textit{VMCBench} with shuffled options.} $M_1$ is PaliGemma-3B, $M_2$ is LLaVA-1.5-7B, $M_3$ is Phi-3.5-Vision, $M_4$ is Qwen2-VL-7B, $M_5$ is Claude-3.5-Sonnet, $M_6$ is GPT-4o.}
    \label{tab:shuffle2}
    \vspace{-1em}
\end{table}

\subsection{Additional Examples from \textit{VMCBench}}

To provide a deeper understanding of \textit{VMCBench}, we include 60 examples (three from each of the 20 sources) in Tables~\ref{tab:examples1} through \ref{tab:examples10}. These examples demonstrate the high quality of our dataset and offer valuable insights for readers.

\input{tables/failures}

\input{tables/vmcbench_full}

\input{tables/prompts}

\input{tables/examples}

%% file: tables/failures.tex
\begin{table*}[!tb]

\rowcolors{2}{white}{light-light-gray}
\setlength\tabcolsep{6pt}
\renewcommand{\arraystretch}{1.1}
\small
\centering

    \begin{tabular}{p{0.1\linewidth}p{0.2\linewidth}p{0.15\linewidth}p{0.15\linewidth}p{0.1\linewidth}p{0.1\linewidth}}
    \toprule
    \textbf{Image} & \textbf{Question} & \textbf{Answer} & \textbf{Prediction} & \textbf{VQAScore} & \textbf{ModelScore} \\
    \midrule

\raisebox{-\totalheight}{\includegraphics[height=5em]{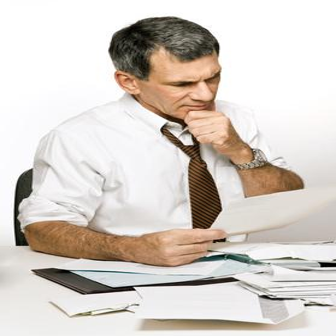}} & Where is the man's left hand holding? & chin & On his chin. & 0 & 1 \\
\raisebox{-\totalheight}{\includegraphics[height=5em]{images/rule2.png}} & How many sinks are there? & 2 & Two sinks. & 0 & 1 \\
\raisebox{-\totalheight}{\includegraphics[height=5em]{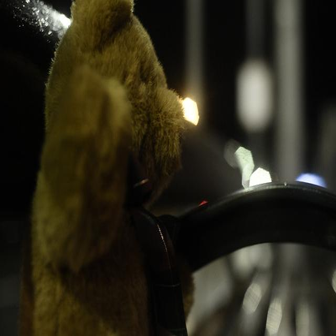}} & What is the light source? & street lights & Streetlights & 0 & 1 \\
\raisebox{-\totalheight}{\includegraphics[height=5em]{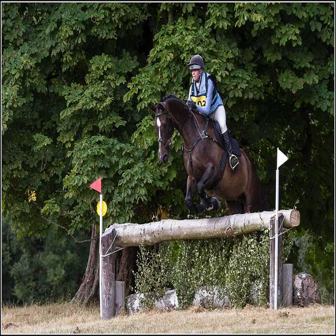}} & What is the horse jumping over? & log & Log fence & 0 & 1 \\
\raisebox{-\totalheight}{\includegraphics[height=5em]{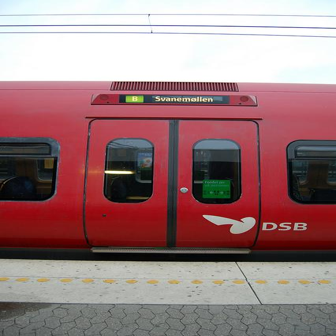}} & Where is the train going? & svanemollen & Svanemøllen & 0 & 1 \\
\raisebox{-\totalheight}{\includegraphics[height=5em]{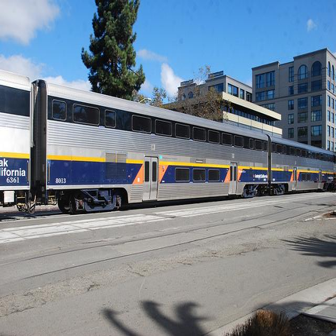}} & What colors are on this train? & blue and yellow and silver & Blue, yellow, and silver. & 0 & 1 \\

    \bottomrule
    \end{tabular}
    \caption{\textbf{Examples of rule-based evaluation failures of open-ended questions.} Rule-based methods fail to account for semantic similarity and penalize formatting errors, resulting in highly inaccurate evaluation results.}
    \label{tab:failures-rule}
\end{table*}

\begin{table*}[!tb]

\rowcolors{2}{white}{light-light-gray}
\setlength\tabcolsep{6pt}
\renewcommand{\arraystretch}{1.1}
\small
\centering

    \begin{tabular}{p{0.1\linewidth}p{0.15\linewidth}p{0.1\linewidth}p{0.25\linewidth}p{0.1\linewidth}p{0.1\linewidth}}
    \toprule
    \textbf{Image} & \textbf{Question} & \textbf{Answer} & \textbf{Prediction} & \textbf{GPT-4o-0513 Score} & \textbf{GPT-4o-0806 Score} \\
    \midrule

\raisebox{-\totalheight}{\includegraphics[height=5em]{images/model1.jpg}} & Does the giraffe appear dangerous? & no & The giraffe in the image does not appear to be dangerous. It seems to be calm and at ease with the person's presence, indicating a level of trust and familiarity between them. & 0.9 & 1.0 \\
\raisebox{-\totalheight}{\includegraphics[height=5em]{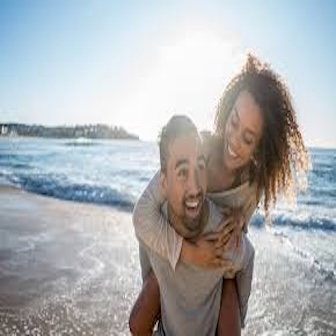}} & Is the man happy? & yes & The man in the image appears to be happy. His facial expression, characterized by a broad smile and eyes that seem to be crinkling at the corners, typically indicates joy or amusement... & 0.9 & 1.0 \\
\raisebox{-\totalheight}{\includegraphics[height=5em]{images/model3.jpg}} & Is that wine in the background? & no & No, there is no wine in the background. The image shows a pizza and a can of Coca-Cola on a desk, with a laptop and some other items in the background. & 0.9 & 1.0 \\
\raisebox{-\totalheight}{\includegraphics[height=5em]{images/model4.jpg}} & What are all the scene text in the image? & S3302CD & The scene text in the image reads 'S 3302 CD'. & 0.9 & 1.0 \\
\raisebox{-\totalheight}{\includegraphics[height=5em]{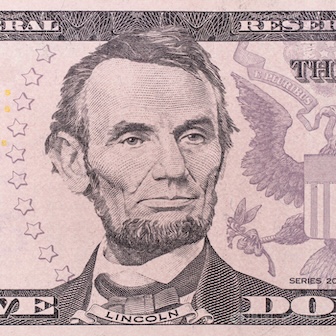}} & What is this banknote? & ten pound or 10 pound or British pound & This is a ten-pound banknote from the United Kingdom, featuring the portrait of Queen Elizabeth II. & 0.9 & 1.0 \\
\raisebox{-\totalheight}{\includegraphics[height=5em]{images/model6.jpg}} & How many birds are there? & 6 & There are six birds perched on the branch, each displaying a unique coloration and size, which suggests they may belong to different species. & 0.9 & 1.0 \\
    \bottomrule
    \end{tabular}
    \caption{\textbf{Examples of model-based evaluation failures of open-ended questions.} GPT-4o-0806 often assigns a perfect score of for similar predictions and answers, whereas GPT-4o-0513 tends to assign a score of 0.9. This behavior variation introduces significant differences in evaluation results and raises concerns about reproducibility in future research. }
    \label{tab:failures-model}
\end{table*}

%% file: tables/vmcbench_full.tex
\begin{table*}[!tb]
\centering
\rowcolors{2}{white}{light-light-gray}
\setlength\tabcolsep{5pt}
\renewcommand{\arraystretch}{1.1}
\scriptsize
\begin{tabular}{lccccccccccc}
\toprule
\textbf{Model} & SEEDBench & MMStar & A-OKVQA & VizWiz & MMVet & VQAv2 & OKVQA & MMMU & MathVista & ScienceQA \\
\midrule
Qwen2-VL-72B & 84.2 & 71.5 & 93.4 & 94.9 & 88.5 & 92.4 & 94.8 & 70.2 & 70.3 & 87.1 \\
GPT-4o & 84.2 & 62.2 & 92.0 & 94.4 & 80.6 & 88.7 & 94.1 & 70.0 & 51.0 & 86.9 \\
Molmo-72B & 82.0 & 62.5 & 88.5 & 88.5 & 79.1 & 85.6 & 94.1 & 59.4 & 60.9 & 89.4 \\
Qwen2-VL-7B & 82.7 & 60.3 & 90.1 & 92.4 & 82.0 & 91.4 & 92.6 & 54.1 & 48.0 & 86.9 \\
Claude-3.5-Sonnet & 78.3 & 53.7 & 86.4 & 87.2 & 87.8 & 84.7 & 91.1 & 59.6 & 56.9 & 79.9 \\
Cambrian-34B & 83.5 & 59.4 & 91.1 & 90.7 & 81.3 & 88.4 & 91.9 & 55.0 & 60.9 & 83.5 \\
Gemini-1.5-Pro & 77.3 & 52.0 & 88.0 & 89.0 & 79.9 & 80.8 & 90.4 & 59.6 & 56.9 & 83.0 \\
VILA1.5-40B & 81.2 & 58.0 & 90.8 & 90.2 & 77.0 & 87.3 & 93.3 & 58.2 & 65.3 & 83.3 \\
GPT-4o-Mini & 79.3 & 47.7 & 86.1 & 92.2 & 80.6 & 88.2 & 92.1 & 56.5 & 43.1 & 78.7 \\
Qwen2-VL-2B & 77.8 & 44.7 & 86.6 & 88.2 & 70.5 & 88.7 & 88.6 & 46.2 & 39.1 & 76.0 \\
CogVLM2-19B & 77.3 & 48.2 & 87.8 & 87.3 & 73.4 & 85.2 & 87.4 & 39.7 & 35.6 & 90.5 \\
Phi-3-Vision & 78.3 & 46.8 & 81.6 & 79.9 & 67.6 & 79.2 & 85.2 & 44.7 & 38.6 & 92.1 \\
Cambrian-13B & 79.3 & 48.5 & 86.4 & 87.3 & 76.3 & 87.0 & 90.6 & 41.3 & 39.6 & 77.4 \\
Cambrian-8B & 78.3 & 53.0 & 84.5 & 88.0 & 68.3 & 85.6 & 87.9 & 43.3 & 45.5 & 77.4 \\
Molmo-7B-D & 74.1 & 46.8 & 82.4 & 81.9 & 63.3 & 80.3 & 83.5 & 43.0 & 37.6 & 91.9 \\
Idefics2-8B & 77.8 & 49.4 & 85.4 & 84.8 & 69.8 & 86.8 & 90.4 & 38.5 & 41.6 & 91.9 \\
Molmo-7B-O & 75.1 & 45.8 & 80.5 & 78.9 & 66.2 & 78.2 & 83.2 & 44.0 & 35.6 & 90.0 \\
Phi-3.5-Vision & 74.8 & 45.8 & 76.5 & 75.7 & 64.0 & 79.4 & 83.5 & 45.2 & 39.1 & 87.3 \\
VILA1.5-13B & 77.5 & 44.9 & 81.9 & 83.3 & 63.3 & 82.4 & 88.6 & 42.8 & 48.5 & 72.6 \\
DeepSeek-VL-7B & 75.6 & 39.9 & 80.5 & 82.1 & 63.3 & 83.6 & 87.7 & 41.1 & 33.2 & 86.7 \\
Molmo-1B & 70.9 & 43.0 & 77.9 & 80.9 & 54.7 & 75.7 & 83.0 & 36.3 & 27.7 & 89.1 \\
CogVLM-17B & 70.9 & 42.8 & 80.7 & 85.0 & 59.7 & 80.1 & 86.7 & 37.5 & 36.1 & 66.7 \\
VILA1.5-8B & 74.3 & 40.9 & 77.9 & 79.4 & 64.7 & 80.8 & 88.6 & 38.0 & 49.0 & 71.5 \\
Gemini-1.5-Flash & 56.3 & 38.0 & 67.0 & 68.5 & 53.2 & 64.1 & 70.6 & 48.1 & 57.9 & 66.3 \\
PaliGemma-3B & 74.6 & 39.9 & 87.3 & 77.0 & 50.4 & 87.7 & 85.2 & 29.1 & 30.7 & 94.3 \\
VILA1.5-3B & 74.3 & 38.5 & 76.9 & 80.9 & 57.6 & 78.5 & 85.4 & 34.4 & 39.6 & 64.9 \\
DeepSeek-VL-1.3B & 70.9 & 37.5 & 74.4 & 82.1 & 52.5 & 80.3 & 84.7 & 31.0 & 22.3 & 63.8 \\
LLaVA1.5-13B & 66.2 & 37.3 & 76.5 & 76.0 & 59.7 & 64.1 & 85.2 & 37.5 & 31.7 & 66.3 \\
LLaVA1.5-7B & 62.2 & 34.2 & 72.5 & 73.8 & 54.0 & 66.7 & 82.2 & 35.6 & 31.2 & 68.6 \\
Chameleon-30B & 53.6 & 33.0 & 57.2 & 52.2 & 48.9 & 58.3 & 68.6 & 34.4 & 32.7 & 57.9 \\
InstructBLIP-7B & 52.8 & 34.7 & 61.9 & 65.4 & 39.6 & 59.7 & 71.9 & 31.0 & 22.8 & 46.8 \\
InstructBLIP-13B & 48.4 & 29.0 & 63.3 & 64.2 & 43.2 & 64.1 & 71.6 & 25.7 & 19.8 & 50.0 \\
Chameleon-7B & 44.9 & 31.6 & 46.4 & 40.4 & 29.5 & 41.4 & 53.1 & 32.7 & 22.3 & 53.6 \\
\midrule
\textbf{Model} & RealWorldQA & GQA & MathVision & TextVQA & OCRVQA & AI2D & ChartQA & DocVQA & InfoVQA & TableVQABench & Avg. \\
\midrule
Qwen2-VL-72B & 78.7 & 91.4 & 38.0 & 98.4 & 95.1 & 85.2 & 90.1 & 99.1 & 92.6 & 83.3 & 85.0 \\
GPT-4o & 75.0 & 84.4 & 34.4 & 97.5 & 95.3 & 81.8 & 80.0 & 98.2 & 79.0 & 76.4 & 80.3 \\
Molmo-72B & 70.0 & 83.1 & 36.9 & 95.3 & 94.0 & 79.7 & 81.2 & 94.7 & 74.9 & 75.2 & 78.7 \\
Qwen2-VL-7B & 67.9 & 89.0 & 30.6 & 97.3 & 95.6 & 75.2 & 78.9 & 98.0 & 78.6 & 69.8 & 78.1 \\
Claude-3.5-Sonnet & 63.2 & 81.9 & 35.1 & 93.6 & 93.2 & 72.9 & 87.9 & 97.7 & 79.1 & 85.6 & 77.8 \\
Cambrian-34B & 65.6 & 87.5 & 42.7 & 94.8 & 96.6 & 74.9 & 78.9 & 89.3 & 64.5 & 59.0 & 77.0 \\
Gemini-1.5-Pro & 60.6 & 79.0 & 49.2 & 91.5 & 93.8 & 72.2 & 72.9 & 83.1 & 64.3 & 70.7 & 74.7 \\
VILA1.5-40B & 63.1 & 88.8 & 33.3 & 91.2 & 95.1 & 71.3 & 73.9 & 77.1 & 57.1 & 57.9 & 74.7 \\
GPT-4o-Mini & 65.8 & 80.2 & 28.3 & 95.1 & 92.5 & 72.7 & 71.6 & 95.3 & 70.5 & 64.0 & 74.0 \\
Qwen2-VL-2B & 57.6 & 83.9 & 32.4 & 94.8 & 91.5 & 68.8 & 70.9 & 95.5 & 65.9 & 61.5 & 71.5 \\
CogVLM2-19B & 56.9 & 81.4 & 29.4 & 96.2 & 88.3 & 68.3 & 75.7 & 92.7 & 62.4 & 63.7 & 71.4 \\
Phi-3-Vision & 54.6 & 79.0 & 29.2 & 92.1 & 89.1 & 69.5 & 78.0 & 93.3 & 61.8 & 66.4 & 70.3 \\
Cambrian-13B & 58.3 & 86.8 & 24.5 & 91.0 & 93.5 & 71.8 & 71.6 & 82.6 & 51.8 & 55.4 & 70.0 \\
Cambrian-8B & 61.5 & 85.3 & 25.2 & 92.4 & 89.6 & 71.3 & 70.9 & 83.1 & 45.4 & 56.5 & 69.6 \\
Molmo-7B-D & 58.7 & 74.1 & 27.9 & 93.0 & 90.4 & 66.3 & 74.8 & 88.9 & 65.7 & 64.9 & 69.5 \\
Idefics2-8B & 55.0 & 80.0 & 27.6 & 91.7 & 93.8 & 66.3 & 67.0 & 81.5 & 47.0 & 47.3 & 68.7 \\
Molmo-7B-O & 55.0 & 75.3 & 26.1 & 90.8 & 86.3 & 67.4 & 74.3 & 84.4 & 58.3 & 59.9 & 67.8 \\
Phi-3.5-Vision & 47.9 & 81.2 & 30.8 & 82.9 & 91.5 & 73.1 & 73.6 & 80.0 & 53.7 & 62.6 & 67.4 \\
VILA1.5-13B & 48.2 & 83.6 & 29.0 & 81.6 & 89.1 & 65.8 & 55.3 & 55.9 & 35.7 & 38.5 & 63.4 \\
DeepSeek-VL-7B & 48.4 & 82.6 & 23.6 & 83.6 & 88.1 & 59.5 & 60.8 & 61.9 & 38.9 & 43.5 & 63.2 \\
Molmo-1B & 45.2 & 76.0 & 26.3 & 87.0 & 87.8 & 60.1 & 62.4 & 79.1 & 51.4 & 48.2 & 63.1 \\
CogVLM-17B & 44.0 & 78.2 & 29.9 & 65.4 & 90.2 & 59.5 & 61.5 & 65.7 & 40.6 & 45.7 & 61.3 \\
VILA1.5-8B & 45.0 & 79.2 & 27.9 & 75.5 & 88.1 & 61.3 & 46.3 & 49.7 & 34.3 & 40.8 & 60.7 \\
Gemini-1.5-Flash & 48.0 & 62.1 & 40.2 & 79.6 & 80.2 & 53.1 & 55.2 & 69.9 & 51.3 & 51.8 & 59.1 \\
PaliGemma-3B & 46.6 & 86.1 & 20.9 & 34.6 & 71.5 & 64.7 & 51.8 & 62.6 & 42.9 & 42.8 & 59.0 \\
VILA1.5-3B & 43.1 & 79.5 & 26.5 & 76.0 & 81.9 & 53.8 & 44.5 & 51.2 & 30.4 & 31.5 & 57.5 \\
DeepSeek-VL-1.3B & 42.4 & 76.5 & 25.6 & 78.9 & 80.1 & 47.6 & 47.7 & 53.9 & 33.6 & 36.3 & 56.1 \\
LLaVA1.5-13B & 34.6 & 81.2 & 25.8 & 67.4 & 84.2 & 56.0 & 29.8 & 38.5 & 30.4 & 30.0 & 53.9 \\
LLaVA1.5-7B & 35.1 & 72.6 & 25.4 & 64.3 & 83.7 & 47.4 & 28.0 & 37.2 & 34.3 & 27.9 & 51.8 \\
Chameleon-30B & 35.3 & 59.2 & 27.6 & 39.3 & 56.7 & 47.2 & 28.4 & 32.1 & 32.0 & 29.7 & 44.2 \\
InstructBLIP-7B & 30.7 & 59.4 & 20.2 & 50.8 & 44.6 & 38.3 & 24.5 & 31.0 & 32.0 & 23.9 & 42.1 \\
InstructBLIP-13B & 26.1 & 60.6 & 25.8 & 48.1 & 49.2 & 33.0 & 20.0 & 28.3 & 24.4 & 26.6 & 41.1 \\
Chameleon-7B & 29.1 & 44.5 & 26.1 & 35.7 & 49.0 & 36.0 & 24.1 & 26.7 & 32.0 & 29.1 & 36.4 \\
\bottomrule
\end{tabular}
\caption{\textbf{Performance of 33 vision language models on 20 subsets of \textit{VMCBench} test set.}}
\label{tab:vlm_results_full}
\end{table*}

%% file: tables/prompts.tex
\lstset{
  basicstyle=\ttfamily,
  keywordstyle=\color{blue},
  commentstyle=\color{green},
  stringstyle=\color{red},
  showstringspaces=false,
  breaklines=true,
  breakatwhitespace=true,
  postbreak=\mbox{\textcolor{red}{$\hookrightarrow$}\space}
}
\begin{figure*}[!tb]
\scriptsize
\begin{lstlisting}
You are an expert in creating challenging and educational multiple-choice questions, specializing in visual interpretation errors.
Your task is to generate plausible but incorrect options (distractors) for given image-based question(s), focusing on misinterpretations of visual information.

Given:
1. One or more images
2. An open-ended question about the image(s)
3. The correct answer to the question

Your task:
1. Carefully analyze and understand the provided image(s). Briefly describe the image content(s) (for your understanding only, do not output this).
2. Generate {num_choice} unique and plausible distractor options based on visual interpretation errors. Each distractor should:
    - Be directly related to misinterpretation of the image(s)
    - Seem potentially correct at first glance
    - Be very misleading for students due to visual misunderstanding
    - Contain a subtle error in interpreting visual information that makes it incorrect
    - Vary in difficulty and the type of visual misinterpretation it represents

3. Ensure you understand how the correct answer relates to specific visual elements in the image(s).
4. Focus on common visual interpretation errors, including:
    - Misreading Graphs or Charts: Create options that misinterpret trends, scales, or relationships in visual data
    - Spatial Misinterpretation: Develop options that misunderstand spatial relationships or perspectives in the image(s)
    - Color Confusion: Include options that misinterpret color-coded information or subtle color differences
    - Pattern Misrecognition: Generate options that incorrectly identify or extend patterns in the image(s)
    - Detail Oversight: Create options that miss crucial details or focus on irrelevant visual elements
    - Scale Misjudgment: Include options that misinterpret the scale or proportions of elements in the image(s)
    - Cross-Image Miscomparison: When multiple images are provided, create options that incorrectly compare or contrast elements across images

5. Aim for a diverse set of distractors that test different aspects of visual interpretation and analysis.
6. Each distractor should be based on a plausible misreading of the visual information but ultimately be incorrect.
7. Consider the specific type(s) of image(s) (e.g., photograph, diagram, graph) and generate errors typical for those visual formats.
8. Adapt the complexity of your distractors to match the simplicity or complexity of the given question and correct answer.
9. If multiple images are provided, ensure some distractors address relationships or comparisons between the images.

10. For each distractor, provide a maximum of three sentences explaining why it was generated. The explanation should describe why this distractor is plausible, the subtle flaw it contains, and how it challenges advanced understanding.

Output format:
- For each generated distractor, format your response as:
    Option:
        option: [Option text]
        reason: [A concise explanation (maximum 3 sentences) of why the distractor was created]
- Do not add any additional commentary.

Remember:
- Your goal is to create challenging yet ultimately incorrect options that specifically target misinterpretations of visual information.
- All distractors should be plausible enough to be considered by a student who hasn't fully developed their visual literacy skills, but clear enough to be definitively incorrect upon careful visual analysis.
- Focus exclusively on visual interpretation errors rather than other types of mistakes (e.g., conceptual misunderstandings, reasoning errors).
- The distractors should directly relate to misunderstandings of the image(s) itself, not just the general topic of the question.
- Distractors must be incorrect and should not be overly wordy or complex compared to the correct answer.
- Ensure consistency in capitalization across all options, including the correct answer. For example, if the correct answer starts with a uppercase letter, adjust all distractors to match.
- When dealing with multiple images, consider how visual interpretation errors might arise from comparing or contrasting information across the images.
- Pay attention to any visual relationships, patterns, or differences that span multiple images, and create distractors that plausibly misinterpret these inter-image connections.
\end{lstlisting}
    \caption{
    \textbf{Detailed prompt for the proposer designed to create distractors addressing \textit{vision} errors.} 
    }
    \label{fig:prompt-visual}
\end{figure*}

\lstset{
  basicstyle=\ttfamily,
  keywordstyle=\color{blue},
  commentstyle=\color{green},
  stringstyle=\color{red},
  showstringspaces=false,
  breaklines=true,
  breakatwhitespace=true,
  postbreak=\mbox{\textcolor{red}{$\hookrightarrow$}\space}
}
\begin{figure*}[!tb]
\scriptsize
\begin{lstlisting}
You are an expert in creating challenging and educational multiple-choice questions, specializing in reasoning errors.
Your task is to generate plausible but incorrect options (distractors) for given image-based question(s), focusing on flaws in logical reasoning and inference.

Given:
1. One or more images
2. An open-ended question about the image(s)
3. The correct answer to the question

Your task:
1. Carefully analyze and understand the provided image(s). Briefly describe the image content(s) (for your understanding only, do not output this).
2. Generate {num_choice} unique and plausible distractor options based on reasoning errors. Each distractor should:
    - Be related to the image(s) and question
    - Seem potentially correct at first glance
    - Be very misleading for students due to faulty reasoning
    - Contain a subtle logical flaw that makes it incorrect
    - Vary in difficulty and the type of reasoning error it represents

3. Ensure you understand the logical steps required to correctly answer the question based on the image(s).
4. Focus on common reasoning errors, including:
    - Complex Reasoning Flaws: Create options that require multi-step reasoning but contain logical gaps or invalid assumptions
    - Causal Inversion: Develop options that reverse cause and effect relationships
    - Context Neglect: Include options that ignore important contextual information provided in the question or image(s)
    - False Analogies: Generate options that draw incorrect parallels or comparisons
    - Hasty Generalizations: Create options that jump to conclusions based on insufficient evidence
    - Cross-Image Fallacies: When multiple images are provided, create options that make invalid logical connections or comparisons between images

5. Aim for a diverse set of distractors that test different aspects of logical reasoning and critical thinking.
6. Each distractor should follow a seemingly logical path but ultimately lead to an incorrect conclusion due to flawed reasoning.
7. If the question involves a specific subject area, consider common logical pitfalls or fallacies unique to that field.
8. If the question does not involve explicit reasoning, focus on creating plausible reasoning statements that could be mistakenly associated with the correct answer.
9. Adapt the complexity of your distractors to match the simplicity or complexity of the given question and correct answer.
10. If multiple images are provided, ensure some distractors address relationships or comparisons between the images, focusing on logical errors in interpreting these relationships.

11. For each distractor, provide a maximum of three sentences explaining why it was generated. The explanation should describe why this distractor is plausible, the subtle flaw it contains, and how it challenges advanced understanding.

Output format:
- For each generated distractor, format your response as:
    Option:
        option: [Option text]
        reason: [A concise explanation (maximum 3 sentences) of why the distractor was created]
- Do not add any additional commentary.

Remember:
- Your goal is to create challenging yet ultimately incorrect options that specifically target flaws in logical reasoning and inference.
- All distractors should be plausible enough to be considered by a student who hasn't fully developed their critical thinking skills, but clear enough to be definitively incorrect upon careful logical analysis.
- Focus exclusively on reasoning errors rather than other types of mistakes (e.g., conceptual misunderstandings, visual misinterpretations).
- Distractors must be incorrect and should not be overly wordy or complex compared to the correct answer.
- Ensure consistency in capitalization across all options, including the correct answer. For example, if the correct answer starts with a uppercase letter, adjust all distractors to match.
- When dealing with multiple images, consider how reasoning errors might arise from comparing or contrasting information across the images.
- Pay attention to any logical relationships, patterns, or differences that span multiple images, and create distractors that plausibly misinterpret these inter-image connections using faulty reasoning.
\end{lstlisting}
    \caption{
\textbf{Detailed prompt for the proposer designed to create distractors addressing \textit{reasoning} errors.} 
    }
    \label{fig:prompt-reasoning}
\end{figure*}

\lstset{
  basicstyle=\ttfamily,
  keywordstyle=\color{blue},
  commentstyle=\color{green},
  stringstyle=\color{red},
  showstringspaces=false,
  breaklines=true,
  breakatwhitespace=true,
  postbreak=\mbox{\textcolor{red}{$\hookrightarrow$}\space}
}
\begin{figure*}[!tb]
\tiny
\begin{lstlisting}
You are an expert in creating challenging and educational multiple-choice questions, specializing in data processing errors.
Your task is to generate plausible but incorrect options (distractors) for given image-based question(s), focusing on mistakes in handling quantitative information and data analysis.

Given:
1. One or more images
2. An open-ended question about the image(s)
3. The correct answer to the question

Your task:
1. Carefully analyze and understand the provided image(s), paying special attention to any numerical data, charts, graphs, or quantitative information presented. Briefly describe the image content(s) (for your understanding only, do not output this).
2. Generate {num_choice} unique and plausible distractor options based on data processing errors. Each distractor should:
    - Be directly related to mishandling of numerical or quantitative information in the image(s)
    - Seem potentially correct at first glance
    - Be very misleading for students due to data processing mistakes
    - Contain a subtle error in calculation, interpretation, or application of quantitative information
    - Vary in difficulty and the type of data processing error it represents

3. Ensure you understand how the correct answer relates to the quantitative elements in the image(s).
4. Focus on common data processing errors, including:
    - Numerical Errors: Create options with incorrect calculations or use of wrong numerical values
    - Unit Conversion Mistakes: Develop options that misapply or neglect unit conversions
    - Statistical Misinterpretation: Include options that misunderstand statistical concepts or misapply statistical tests
    - Data Range Errors: Generate options that incorrectly interpret data ranges or outliers
    - Temporal/Sequential Errors: Create options with mistakes in the order or timing of data points or processes
    - Correlation/Causation Confusion: Include options that mistake correlation for causation in data relationships
    - Sampling Errors: Develop options that misinterpret sample sizes or sampling methods
    - Rounding Errors: Create options with incorrect rounding or significant figure usage

5. Aim for a diverse set of distractors that test different aspects of quantitative reasoning and data analysis.
6. Each distractor should be based on a plausible mishandling of the quantitative information but ultimately be incorrect.
7. Consider the specific type of data presented (e.g., discrete vs. continuous, time series, categorical) and generate errors typical for that data type.
8. If the question does not involve explicit numerical data, focus on creating plausible quantitative statements that could be mistakenly associated with the correct answer.
9. Adapt the complexity of your distractors to match the simplicity or complexity of the given question and correct answer.
10. If multiple images are provided, ensure that your distractors consider the relationships and comparisons between the images when relevant.
11. When generating numerical distractors:
    - Carefully analyze the structure and precision of the correct answer
    - Create distractors that closely mimic the format, precision, and magnitude of the correct answer
    - Use a mix of common calculation errors, transposition mistakes, and misinterpretations to generate deceptive options
    - For answers with specific formats (e.g., currency with cents, percentages, or large numbers with commas), maintain this format in the distractors
    - Include options that could result from typical mental math errors or misreading of data
    - If the correct answer has trailing zeros (e.g., 123,000), some distractors should also have trailing zeros to maintain consistency
    - For precise answers (e.g., $493.02), create distractors with same precision (e.g., $439.20, $493.20, $492.03) to increase difficulty while maintaining consistency in decimal places
12. Ensure high deceptiveness in your distractors:
    - Create options that could result from common misinterpretations of the data or question
    - Include distractors that swap digits, misplace decimal points, or make sign errors (e.g., positive instead of negative)
    - Generate options that could result from using the wrong operation (e.g., addition instead of subtraction)
    - For multi-step calculations, include results that would occur if a step was omitted or performed incorrectly
    - Consider psychological factors that might lead to specific errors, such as anchoring bias or confirmation bias
13. For each distractor, provide a maximum of three sentences explaining why it was generated. The explanation should describe why this distractor is plausible, the subtle flaw it contains, and how it challenges advanced understanding.

Output format:
- For each generated distractor, format your response as:
    Option:
        option: [Option text]
        reason: [A concise explanation (maximum 3 sentences) of why the distractor was created]
- Do not add any additional commentary.

Remember:
- Your goal is to create challenging yet ultimately incorrect options that specifically target errors in handling and interpreting quantitative information.
- All distractors should be plausible enough to be considered by a student who hasn't fully developed their quantitative reasoning skills, but clear enough to be definitively incorrect upon careful analysis.
- Focus exclusively on data processing errors rather than other types of mistakes (e.g., conceptual misunderstandings, visual misinterpretations).
- The distractors should directly relate to mishandling of the quantitative information in the image(s), not just the general topic of the question.
- Ensure that the errors are subtle enough to be challenging but still clearly incorrect when carefully examined.
- Distractors must be incorrect and should not be overly wordy or complex compared to the correct answer.
- Ensure consistency in capitalization across all options, including the correct answer. For example, if the correct answer starts with a uppercase letter, adjust all distractors to match.
- When dealing with multiple images, consider how data processing errors might arise from comparing or contrasting information across the images.
- Pay attention to any relationships, trends, or patterns that span multiple images, and create distractors that plausibly misinterpret these inter-image connections.
\end{lstlisting}
    \caption{
\textbf{Detailed prompt for the proposer designed to create distractors addressing \textit{data processing} errors.} 
    }
    \label{fig:prompt-data}
\end{figure*}

\lstset{
  basicstyle=\ttfamily,
  keywordstyle=\color{blue},
  commentstyle=\color{green},
  stringstyle=\color{red},
  showstringspaces=false,
  breaklines=true,
  breakatwhitespace=true,
  postbreak=\mbox{\textcolor{red}{$\hookrightarrow$}\space}
}
\begin{figure*}[!tb]
\scriptsize
\begin{lstlisting}
You are an expert in creating challenging and educational multiple-choice questions, specializing in conceptual errors.
Your task is to generate plausible but incorrect options (distractors) for given image-based question(s), focusing on conceptual misunderstandings and misconceptions.

Given:
1. One or more images
2. An open-ended question about the image(s)
3. The correct answer to the question

Your task:
1. Carefully analyze and understand the provided image(s). Briefly describe the image content(s) (for your understanding only, do not output this).
2. Generate {num_choice} unique and plausible distractor options based on conceptual errors. Each distractor should:
    - Be related to the image(s) and question
    - Seem potentially correct at first glance
    - Be very misleading for students due to conceptual misunderstandings
    - Contain a subtle flaw or misconception that makes it incorrect
    - Vary in difficulty and the type of conceptual error it represents

3. Ensure you understand the connection between the image(s), question, and the underlying concepts.
4. Focus on common conceptual misconceptions in the subject area, including:
    - Concept Confusion: Create options that are similar to the correct concept but with subtle differences
    - Partial Correctness: Include options that contain partially correct information but are incomplete or misleading
    - Overgeneralization: Develop options that incorrectly apply specific cases to general situations
    - Cross-Image Misconceptions: When multiple images are provided, create options that misapply concepts across different images

5. Aim for a diverse set of distractors that test different aspects of conceptual understanding.
6. Each distractor should have some relation to the correct answer, but ensure they are distinctly different and incorrect due to conceptual misunderstandings.
7. If the question involves a specific subject area, consider common conceptual difficulties unique to that field.
8. Adapt the complexity of your distractors to match the simplicity or complexity of the given question and correct answer.
9. If multiple images are provided, ensure some distractors address relationships or comparisons between the images, focusing on conceptual errors in interpreting these relationships.

10. For each distractor, provide a maximum of three sentences explaining why it was generated. The explanation should describe why this distractor is plausible, the subtle flaw it contains, and how it challenges advanced understanding.

Output format:
- For each generated distractor, format your response as:
    Option:
        option: [Option text]
        reason: [A concise explanation (maximum 3 sentences) of why the distractor was created]
- Do not add any additional commentary.

Remember:
- Your goal is to create challenging yet ultimately incorrect options that specifically target conceptual misunderstandings.
- All distractors should be plausible enough to be considered by a student who doesn't fully grasp the concept, but clear enough to be definitively incorrect upon careful consideration.
- Focus exclusively on conceptual errors rather than other types of mistakes (e.g., calculation errors, visual misinterpretations).
- Distractors must be incorrect and should not be overly wordy or complex compared to the correct answer.
- Ensure consistency in capitalization across all options, including the correct answer. For example, if the correct answer starts with a uppercase letter, adjust all distractors to match.
- When dealing with multiple images, consider how conceptual errors might arise from comparing or contrasting information across the images.
- Pay attention to any conceptual relationships, patterns, or differences that span multiple images, and create distractors that plausibly misinterpret these inter-image connections due to conceptual misunderstandings.
\end{lstlisting}
    \caption{
\textbf{Detailed prompt for the proposer designed to create distractors addressing \textit{concept} errors.} 
    }
    \label{fig:prompt-concept}
\end{figure*}

\lstset{
  basicstyle=\ttfamily,
  keywordstyle=\color{blue},
  commentstyle=\color{green},
  stringstyle=\color{red},
  showstringspaces=false,
  breaklines=true,
  breakatwhitespace=true,
  postbreak=\mbox{\textcolor{red}{$\hookrightarrow$}\space}
}
\begin{figure*}[!tb]
\scriptsize
\begin{lstlisting}
You are an expert in creating extremely challenging multiple-choice questions, specializing in highly sophisticated question-focused distractors.

Your task is to generate plausible but incorrect options (distractors) for given questions, focusing on creating the most difficult and deceptive answers based on the question text.

Given:
1. An open-ended question
2. The correct answer to the question

Your task:
1. Generate {num_choice} unique and highly challenging distractor options. Each distractor should:
- Be closely related to the question text
- Seem very plausible and potentially correct even upon careful consideration
- Be extremely misleading, requiring deep understanding to recognize as incorrect
- Contain subtle, sophisticated flaws that make them incorrect
- Represent the highest level of difficulty and complexity

2. Focus on creating distractors that:
- Leverage advanced knowledge or nuanced interpretations of the subject matter
- Provide logically sound but ultimately incorrect answers based on the question
- Exploit common high-level misconceptions or advanced misinterpretations
- Offer highly plausible alternatives that might be true in many situations but are incorrect in this specific context

3. Aim for a diverse set of sophisticated distractors that challenge different aspects of advanced understanding and critical thinking.

4. Each distractor should be intricately related to the question topic and the correct answer, but with crucial differences that make them incorrect.

5. If the question involves a specific subject area, incorporate advanced concepts and potential misunderstandings at an expert level.

6. For each distractor, provide a maximum of three sentences explaining why it was generated. The explanation should describe why this distractor is plausible, the subtle flaw it contains, and how it challenges advanced understanding.

Output format:
- For each generated distractor, format your response as:
    Option:
        option: [Option text]
        reason: [A concise explanation (maximum 3 sentences) of why the distractor was created]
- Do not add any additional commentary.

Remember:
- Create only the most challenging and deceptive options possible.
- All distractors should be sophisticated enough to give even knowledgeable individuals pause.
- Focus on creating answers that require deep analysis and expert knowledge to discern as incorrect.
- Ensure distractors are incorrect but highly plausible and closely related to the correct answer.
- Maintain consistency in style, complexity, and structure across all options, matching the correct answer's sophistication.
- Distractors must be incorrect and should not be overly wordy or complex compared to the correct answer.
\end{lstlisting}
    \caption{
    \textbf{\textbf{Detailed prompt for the proposer designed to create distractors addressing \textit{question bias} errors.} } 
    }
    \label{fig:prompt-bias}
\end{figure*}

\lstset{
  basicstyle=\ttfamily,
  keywordstyle=\color{blue},
  commentstyle=\color{green},
  stringstyle=\color{red},
  showstringspaces=false,
  breaklines=true,
  breakatwhitespace=true,
  postbreak=\mbox{\textcolor{red}{$\hookrightarrow$}\space}
}
\begin{figure*}[!tb]
\scriptsize
\begin{lstlisting}
Task: Analyze and enhance the provided distractors, which were generated based on {type} error type, to maximize their difficulty and deceptiveness while ensuring they remain incorrect.

Given:
1. One or more images
2. A question about the image(s)
3. The correct answer
4. A set of distractor options for a specific error type (e.g., reasoning error, question bias, etc.)
5. The reasoning provided for why each distractor was created

For each distractor, your task is to:
1. Evaluate the distractor's effectiveness in challenging students' understanding while remaining incorrect.
2. Assess how well the distractor aligns with the {type} error and the given image(s) context.
3. Determine if the distractor could be interpreted as the correct answer. If so, add suggestions towards this.
4. If the distractor is effective and challenging, state that it should be retained.
5. If improvements are needed, provide specific suggestions to increase the distractor's difficulty and deceptiveness without:
    a. Increasing the option's length or adding unnecessary modifiers
    b. Making the distractor correct
6. Ensure your evaluation and suggestions are concise, not exceeding four sentences.

Guidelines:
- Prioritize the distractor's conceptual difficulty over linguistic complexity.
- If a distractor is correct or could be interpreted as correct, clearly state this and suggest how to modify it to make it unambiguously incorrect.
- Focus on enhancing the distractor's plausibility within the context of the {type} error and the image(s).
- Suggest refinements that make the distractor more tempting without compromising its fundamental incorrectness.
- Ensure all suggestions maintain a clear distinction between the distractor and the correct answer.

For each option, format your response as:
Option:
    option: [Option text]
    comment: [Your evaluation and specific suggestions, if needed, or confirmation of effectiveness]
\end{lstlisting}
    \caption{
\textbf{Detailed prompt for the \textit{reviewer}, whose feedback iteratively refines the distractors to improve their quality.} 
    }
    \label{fig:prompt-reviewer}
\end{figure*}

\lstset{
  basicstyle=\ttfamily,
  keywordstyle=\color{blue},
  commentstyle=\color{green},
  stringstyle=\color{red},
  showstringspaces=false,
  breaklines=true,
  breakatwhitespace=true,
  postbreak=\mbox{\textcolor{red}{$\hookrightarrow$}\space}
}
\begin{figure*}[!tb]
\scriptsize
\begin{lstlisting}
You are an expert Selection Agent tasked with curating the most challenging and high-quality distractor options for multiple-choice questions based on one or more provided images.

Your goal is to select the best {fusion_selected_choice_num} unique distractors from a pool of multiple distractors, ensuring a diverse, non-repetitive, and challenging set of options that are relevant to the given image(s).
Given:
- One or more images related to the question
- A dictionary containing multiple distractor options, organized into five categories:
    1. Concept Error ({num_choice} options)
    2. Reasoning Error ({num_choice} options)
    3. Visual Interpretation Error ({num_choice} options)
    4. Data Processing Error ({num_choice} options)
    5. Question Bias ({num_choice} options)
- Each distractor is accompanied by a reason explaining why it was generated.

Your task:
1. Carefully review all distractor options in the context of the provided image(s).
2. Select the top {fusion_selected_choice_num} distractors based on the following criteria:
- Image relevance: Prioritize distractors that are closely related to the content, context, or details present in the given image(s).
- Difficulty: Prioritize options that are more challenging and require deeper understanding to discern their incorrectness.
- Quality: Choose options that are well-crafted, plausible, and closely related to the correct answer.
- Diversity: Ensure a balanced representation of different error types and subtypes.
- Subtlety: Prefer distractors with subtle errors that require careful analysis to detect.
- Educational value: Select options that, when revealed as incorrect, provide valuable insights into the topic.
- Uniqueness: Ensure that each selected distractor is distinct from others in meaning and approach, avoiding repetition or highly similar concepts.
- Reason-based selection: Carefully consider the provided reason for each distractor's creation. Prioritize distractors whose reasoning aligns well with the image context, question intent, or presents a strong challenge for test-takers. Use the quality of these reasons to guide your selection process.

3. Ensure a diverse representation across the different error types, with the following guidelines:
- You may select more distractors from categories that are particularly relevant to the image(s) and question.
- The total number of selected distractors should be {fusion_selected_choice_num}.
4. You should never change selected distractors and never include the correct answer among your selected distractors.

Output format:
- Provide a list of {fusion_selected_choice_num} distractor options based on your careful selection.
- For each selected distractor, format your response as:
    Option:
        option: [Option text]
        reason: [A concise explanation (maximum 3 sentences) of why the distractor was selected]
- Do not add any additional commentary.

Remember:
- Your primary goal is to create a challenging yet educational set of distractors that will effectively test students' understanding of the subject matter in relation to the provided image(s).
- If the given correct answer is a list, ensure that none of the selected distractors are included in the correct answer.
- Ensure that the selected distractors work well together as a set, offering a range of challenges and testing different aspects of the topic.
- Consider how each distractor might interact with the others and with the correct answer to create a cohesive and challenging question.
- Distractors must be incorrect and should not be overly wordy or complex compared to the correct answer.
- Ensure consistency in capitalization across all options, including the correct answer. If the correct answer begins with a uppercase letter, adjust all distractors to match.
- Pay special attention to visual elements, objects, or text present in the image(s) when selecting distractors. Incorporate these image-based elements into your selections when relevant.
- If multiple images are provided, ensure that the selected distractors are relevant across all images or specifically address the relationships between the images.
- Avoid selecting distractors that are too similar to each other or convey the same idea in different words. 
\end{lstlisting}
    \caption{
\textbf{Detailed prompt for the \textit{selector}, which guides the selection of the three most challenging distractors to enhance question difficulty.} 
    }
    \label{fig:prompt-selector}
\end{figure*}

\lstset{
  basicstyle=\ttfamily,
  keywordstyle=\color{blue},
  commentstyle=\color{green},
  stringstyle=\color{red},
  showstringspaces=false,
  breaklines=true,
  breakatwhitespace=true,
  postbreak=\mbox{\textcolor{red}{$\hookrightarrow$}\space}
}
\begin{figure*}[!tb]
\scriptsize
\begin{lstlisting}
Your task is to evaluate a multiple-choice question (with accompanying image) to determine if any incorrect choices (distractors) could also be considered correct answers.

CRITICAL: The marked correct answer MUST always be treated as valid and correct, regardless of your own assessment. Never question or evaluate the correct answer - your task is to accept it as an absolute truth and evaluate only whether other choices could also be correct.

Score the question's correctness using this scale:
5 - Perfect: All other choices are clearly incorrect
4 - Good: Other choices are mostly wrong but have minor elements of correctness
3 - Fair: At least one other choice could be partially correct
2 - Poor: At least one other choice could be equally correct
1 - Invalid: Multiple choices are equally valid as the correct answer

Provide:
1. Score (1-5)
2. Brief explanation focusing specifically on any problematic distractor choices
3. Suggested improvements for the problematic distractors (if applicable)

Remember: Never analyze whether the marked correct answer is right or wrong - it is ALWAYS correct by definition. Focus exclusively on whether other choices could also be valid answers.
\end{lstlisting}
    \caption{
\textbf{Detailed prompt for the \textit{evaluator}, which evaluates the correctness of the generated questions, defined as there is only one correct answer.} 
    }
    \label{fig:prompt-evaluator}
\end{figure*}

\lstset{
  basicstyle=\ttfamily,
  keywordstyle=\color{blue},
  commentstyle=\color{green},
  stringstyle=\color{red},
  showstringspaces=false,
  breaklines=true,
  breakatwhitespace=true,
  postbreak=\mbox{\textcolor{red}{$\hookrightarrow$}\space}
}
\begin{figure*}[!tb]
\scriptsize
\begin{lstlisting}
You are an expert in educational assessment design specializing in multiple-choice question improvement. Your task is to enhance question effectiveness by revising problematic distractors (incorrect answer choices) while maintaining the existing correct answer.

Input Required:
1. The complete question
2. The current correct answer
3. Any associated images/materials
4. Specific feedback about problematic distractors
5. Suggested improvements (if provided)

Analysis Steps:
1. Review the question content and learning objective
2. Analyze the designated correct answer
3. Examine the feedback regarding problematic distractors
4. Evaluate any provided suggestions for improvement:
- Assess if suggestions fully address the identified issues
- Determine if suggestions align with best practices
- Identify any gaps or weaknesses in the suggestions
5. Develop exactly 3 improved distractors that:
- Are plausible but clearly incorrect
- Address the identified issues
- Align with common student misconceptions
- Maintain consistent format and length with other options
- Go beyond provided suggestions when necessary for better quality

Guidelines:
1. Treat the marked correct answer as fixed and unchangeable
2. Only modify distractors specifically identified as problematic
3. Preserve any well-functioning distractors
4. Maintain the original difficulty level of the question
5. Use your expertise to improve upon or deviate from provided suggestions if they:
- Are too vague or incomplete
- Don't fully address the identified issues
- Could be enhanced for better assessment quality
- Miss important misconceptions or learning opportunities

Output:
1. Brief analysis of the distractor issues and improvement approach
2. Three improved distractors
\end{lstlisting}
    \caption{
\textbf{Detailed prompt for the \textit{refiner}, which ensures the correctness of the generated questions, guaranteeing that there is only one correct answer.} 
    }
    \label{fig:prompt-refiner}
\end{figure*}

%% file: tables/examples.tex
\begin{table*}[!tb]

\rowcolors{2}{white}{light-light-gray}
\setlength\tabcolsep{6pt}
\renewcommand{\arraystretch}{1.1}
\small
\centering

    \begin{tabular}{p{0.1\linewidth}p{0.1\linewidth}p{0.3\linewidth}p{0.4\linewidth}}
    \toprule
    \textbf{Source} & \textbf{Image} & \textbf{Question} & \textbf{Choices} \\
    \midrule
A-OKVQA & \raisebox{-\totalheight}{\includegraphics[height=5em]{images/8255.jpg}} & What season of the year is shown here? & A. late summer with green leaves \newline B. early spring with blooming flowers \newline \textcolor{orange}{C. fall} \newline D. early winter with snow \\
A-OKVQA & \raisebox{-\totalheight}{\includegraphics[height=5em]{images/7998.jpg}} & What occasion are the bears probably sitting at the table enjoying? & A. Thanksgiving \newline B. Easter \newline C. New Year's Eve \newline \textcolor{orange}{D. Christmas} \\
A-OKVQA & \raisebox{-\totalheight}{\includegraphics[height=5em]{images/8430.jpg}} & What kind of beverage is the red sign advertising? & A. Dr Pepper \newline \textcolor{orange}{B. Coca Cola} \newline C. Red Bull \newline D. Pepsi \\
AI2D & \raisebox{-\totalheight}{\includegraphics[height=5em]{images/1667.jpg}} & Which shows the first stage? & A. b \newline B. a \newline C. d \newline \textcolor{orange}{D. c} \\
AI2D & \raisebox{-\totalheight}{\includegraphics[height=5em]{images/1293.jpg}} & What part of plants the diagram depicts? & \textcolor{orange}{A. Leaf} \newline B. Stem \newline C. Root \newline D. Flower petal \\
AI2D & \raisebox{-\totalheight}{\includegraphics[height=5em]{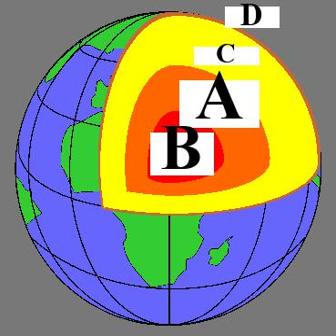}} & Which is the exterior portion of the earth? & A. A \newline \textcolor{orange}{B. D} \newline C. C \newline D. B \\
ChartQA & \raisebox{-\totalheight}{\includegraphics[height=5em]{images/7120.jpg}} & What percentage of respondents own lots of vinyl records? & A. 35 \newline \textcolor{orange}{B. 24} \newline C. 30 \newline D. 28 \\
ChartQA & \raisebox{-\totalheight}{\includegraphics[height=5em]{images/7076.jpg}} & In which year is the ACSI score is lowest? & \textcolor{orange}{A. 2015} \newline B. 2017 \newline C. 2018 \newline D. 2010 \\
ChartQA & \raisebox{-\totalheight}{\includegraphics[height=5em]{images/7073.jpg}} & What is the total ratio of 2014 through 2017? & A. 5.36 \newline \textcolor{orange}{B. 5.11} \newline C. 4.57 \newline D. 5.25 \\
    \bottomrule
    \end{tabular}
    \caption{\textbf{Examples of \textit{VMCBench} (1/7).} Each example has a dataset source, image, question, and four choices (correct choice highlighted in \textcolor{orange}{orange}).}
    \label{tab:examples1}
\end{table*}

\begin{table*}[!tb]

\rowcolors{2}{white}{light-light-gray}
\setlength\tabcolsep{6pt}
\renewcommand{\arraystretch}{1.1}
\small
\centering

    \begin{tabular}{p{0.1\linewidth}p{0.1\linewidth}p{0.3\linewidth}p{0.4\linewidth}}
    \toprule
    \textbf{Source} & \textbf{Image} & \textbf{Question} & \textbf{Choices} \\
    \midrule

DocVQA & \raisebox{-\totalheight}{\includegraphics[height=5em]{images/5929.jpg}} & What is the table number? & \textcolor{orange}{A. 7} \newline B. 8 \newline C. 10 \newline D. 9 \\
DocVQA & \raisebox{-\totalheight}{\includegraphics[height=5em]{images/5663.jpg}} & In the plot, what is the value of "r"? & \textcolor{orange}{A. 0.994} \newline B. 0.949 \newline C. 1.004 \newline D. 0.980 \\
DocVQA & \raisebox{-\totalheight}{\includegraphics[height=5em]{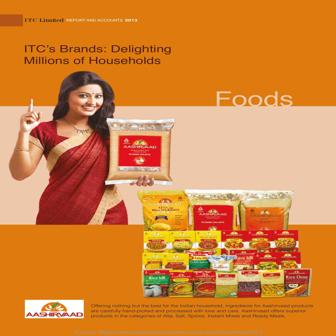}} & Which category item's advertisement is this? & A. health supplements \newline \textcolor{orange}{B. foods} \newline C. home appliances \newline D. clothing \\
GQA & \raisebox{-\totalheight}{\includegraphics[height=5em]{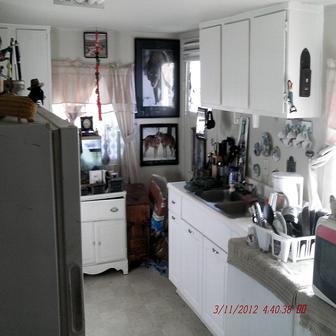}} & What food is to the left of the table? & A. can of soup \newline B. bag of flour \newline C. cereal box \newline \textcolor{orange}{D. dog food} \\
GQA & \raisebox{-\totalheight}{\includegraphics[height=5em]{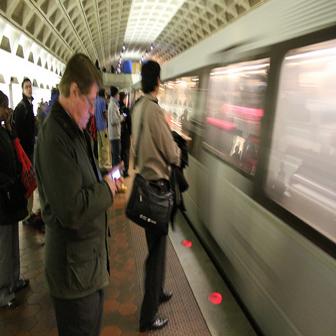}} & Who is looking at the cell phone? & \textcolor{orange}{A. man} \newline B. tourist \newline C. child \newline D. nobody \\
GQA & \raisebox{-\totalheight}{\includegraphics[height=5em]{images/7776.jpg}} & What does the woman hold? & \textcolor{orange}{A. cell phone} \newline B. digital camera \newline C. remote control \newline D. compact mirror \\
InfoVQA & \raisebox{-\totalheight}{\includegraphics[height=5em]{images/5343.jpg}} & Which states/UT has been included under ``Certain"  risk of community transmission? & A. manipur, tamil nadu \newline B. maharashtra, gujarat \newline C. rajasthan, karnataka \newline \textcolor{orange}{D. telangana, delhi} \\
InfoVQA & \raisebox{-\totalheight}{\includegraphics[height=5em]{images/4993.jpg}} & What percentage of Canadian still go to brick and mortar stores to buy items? & A. 30\% \newline \textcolor{orange}{B. 70\%} \newline C. 80\% \newline D. 60\% \\
InfoVQA & \raisebox{-\totalheight}{\includegraphics[height=5em]{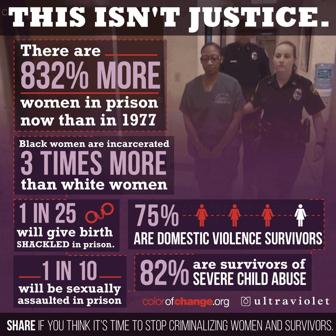}} & How many women out of every 4 women are domestic violence survivors? & A. 4 \newline \textcolor{orange}{B. 3} \newline C. 1.5 \newline D. 2 \\
    \bottomrule
    \end{tabular}
    \caption{\textbf{Examples of \textit{VMCBench} (2/7).} Each example has a dataset source, image, question, and four choices (correct choice highlighted in \textcolor{orange}{orange}).}
    \label{tab:examples2}
\end{table*}

\begin{table*}[!tb]

\rowcolors{2}{white}{light-light-gray}
\setlength\tabcolsep{6pt}
\renewcommand{\arraystretch}{1.1}
\small
\centering

    \begin{tabular}{p{0.1\linewidth}p{0.1\linewidth}p{0.41\linewidth}p{0.3\linewidth}}
    \toprule
    \textbf{Source} & \textbf{Image} & \textbf{Question} & \textbf{Choices} \\
    \midrule

MMMU & \raisebox{-\totalheight}{\includegraphics[height=5em]{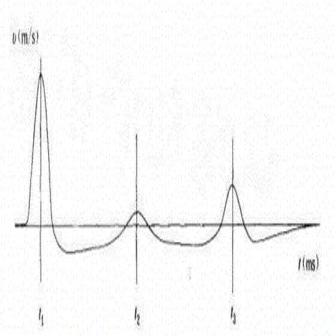}} & The average wave velocity value of the bored pile body at a certain site is 3555.6m/s, and the low strain reflected wave dynamic test curve of a certain column is shown in Figure 10-3, corresponding to the values of time t1, t2, and t3 in the figure, which of the following options is the closest value to the length of the pile?  & A. 22.0m \newline \textcolor{orange}{B. 24.0m} \newline C. 26.0m \newline D. 23.0m \\
MMMU & \raisebox{-\totalheight}{\includegraphics[height=5em]{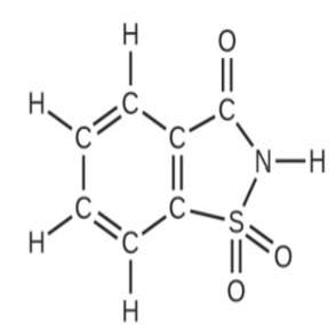}} & How many molecules of the sweetener saccharin can be prepared from 30 $C$ atoms, 25 $H$ atoms, 12 $O$ atoms, 8 $S$ atoms, and 14 $N$ atoms? & A. 6 \newline \textcolor{orange}{B. 4} \newline C. 2 \newline D. 5 \\
MMMU & \raisebox{-\totalheight}{\includegraphics[height=5em]{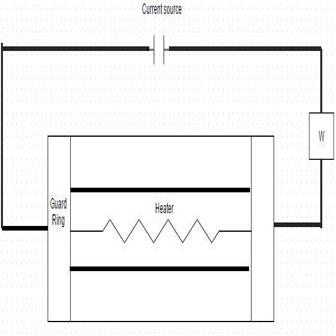}} & The accompanying sketch shows the schematic arrangement for measuring the thermal conductivity by the guarded hot plate method. Two similar 1 cm thick specimens receive heat from a 6.5 cm by 6.5 cm guard heater. When the power dissipation by the wattmeter was 15 W, the thermocouples inserted at the hot and cold surfaces indicated temperatures as 325 K and 300 K. What is the thermal conductivity of the test specimen material? & A. 0.86 W/m K \newline B. 0.5 W/m K \newline C. 0.68 W/m K \newline \textcolor{orange}{D. 0.71 W/m K} \\
MMStar & \raisebox{-\totalheight}{\includegraphics[height=5em]{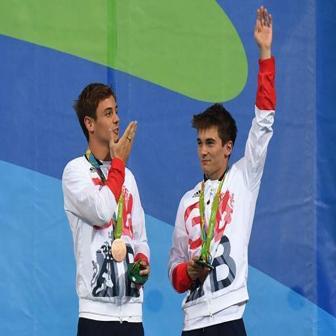}} & What color is the ribbon that the man on the right is holding? & A. Red \newline B. Blue \newline C. Yellow \newline \textcolor{orange}{D. Green} \\
MMStar & \raisebox{-\totalheight}{\includegraphics[height=5em]{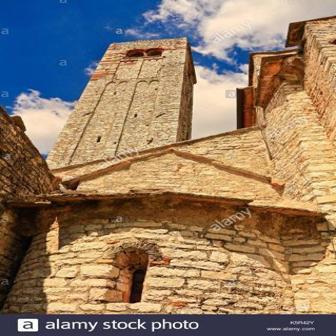}} & What is the main feature of the building in the image? & A. The colorful facade \newline B. The large stained glass windows \newline C. The marble columns \newline \textcolor{orange}{D. The stone wall} \\
MMStar & \raisebox{-\totalheight}{\includegraphics[height=5em]{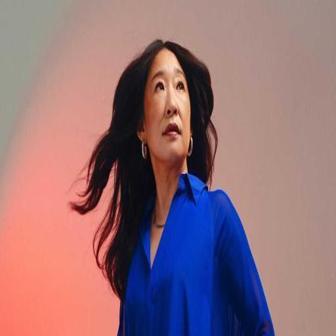}} & who is this person? & A. Awkwafina \newline \textcolor{orange}{B. Sandra Oh} \newline C. Ali Wong \newline D. Lucy Liu \\
MMVet & \raisebox{-\totalheight}{\includegraphics[height=5em]{images/2784.jpg}} & What is the original price for pork belly before discount? & A. 10 \newline B. 12 \newline \textcolor{orange}{C. 14} \newline D. 15 \\
MMVet & \raisebox{-\totalheight}{\includegraphics[height=5em]{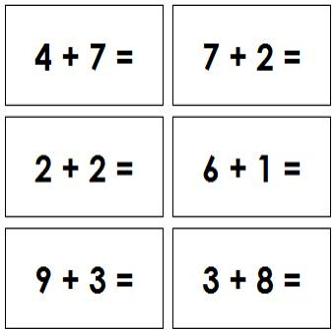}} & What is the answer to the second equation on the right? & A. 11 \newline B. 5 \newline \textcolor{orange}{C. 7} \newline D. 9 \\
MMVet & \raisebox{-\totalheight}{\includegraphics[height=5em]{images/2902.jpg}} & What occasions would someone use this meme? & \textcolor{orange}{A. Sharing relatable humor about feeling sleepy or having conflicting desires, especially during the day.} \newline B. Expressing excitement about a new bedtime routine \newline C. Celebrating an all-nighter successfully pulled off \newline D. Promoting productivity in the workplace \\

    \bottomrule
    \end{tabular}
    \caption{\textbf{Examples of \textit{VMCBench} (3/7).} Each example has a dataset source, image, question, and four choices (correct choice highlighted in \textcolor{orange}{orange}).}
    \label{tab:examples4}
\end{table*}

\begin{table*}[!tb]

\rowcolors{2}{white}{light-light-gray}
\setlength\tabcolsep{6pt}
\renewcommand{\arraystretch}{1.1}
\small
\centering

    \begin{tabular}{p{0.1\linewidth}p{0.1\linewidth}p{0.3\linewidth}p{0.4\linewidth}}
    \toprule
    \textbf{Source} & \textbf{Image} & \textbf{Question} & \textbf{Choices} \\
    \midrule

MathVision & \raisebox{-\totalheight}{\includegraphics[height=5em]{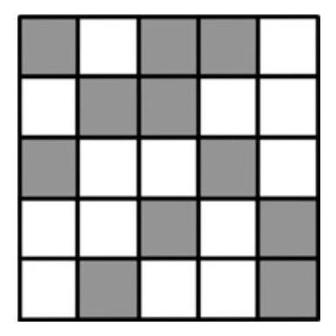}} & In the grid, how many grey squares have to be coloured white, so that in each row and each column there is exactly one grey square? & A. 5 \newline B. 8 \newline \textcolor{orange}{C. 6} \newline D. 7 \\
MathVision & \raisebox{-\totalheight}{\includegraphics[height=5em]{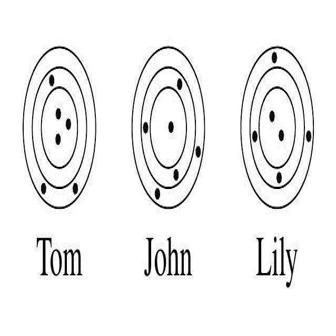}} & Tom, John and Lily each shot six arrows at a target. Arrows hitting anywhere within the same ring scored the same number of points. Tom scored 46 points and John scored 34 points, as shown. How many points did Lily score?  & A. 42 \newline \textcolor{orange}{B. 40} \newline C. 44 \newline D. 38 \\
MathVision & \raisebox{-\totalheight}{\includegraphics[height=5em]{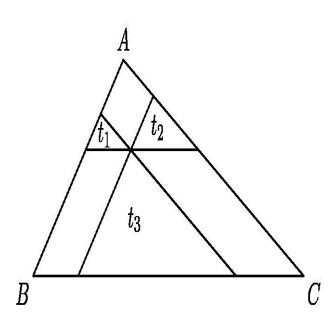}} & A point $P$ is chosen in the interior of $\triangle ABC$ so that when lines are drawn through $P$ parallel to the sides of $\triangle ABC$, the resulting smaller triangles, $t_1$, $t_2$, and $t_3$ in the figure, have areas 4, 9, and 49, respectively. Find the area of $\triangle ABC$. & A. 81 \newline B. 128 \newline \textcolor{orange}{C. 144} \newline D. 72 \\
MathVista & \raisebox{-\totalheight}{\includegraphics[height=5em]{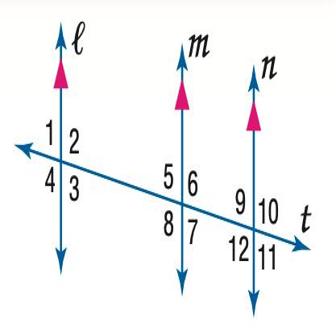}} & In the figure, $\angle 9 = 75$. Find the measure of $\angle 6$. & A. 120 \newline B. 135 \newline C. 150 \newline \textcolor{orange}{D. 105} \\
MathVista & \raisebox{-\totalheight}{\includegraphics[height=5em]{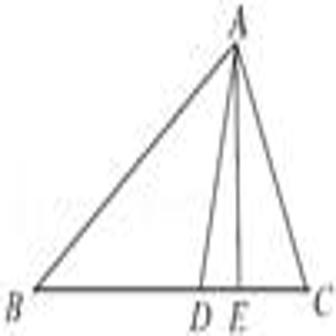}} & (Original in Chinese) As shown in the figure, in \( \triangle ABC \), \( AD \) is the angle bisector, and \( AE \) is the altitude. If \( \angle B = 40^\circ \) and \( \angle C = 70^\circ \), then the measure of \( \angle EAD \) is (). & A. 25° \newline B. 20° \newline C. 30° \newline \textcolor{orange}{D. 15°} \\
MathVista & \raisebox{-\totalheight}{\includegraphics[height=5em]{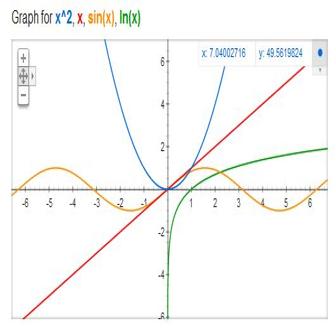}} & What is the green curve? & A. a cubic function \newline B. a trigonometric function \newline C. an exponential function \newline \textcolor{orange}{D. a logarithmic function} \\
OCRVQA & \raisebox{-\totalheight}{\includegraphics[height=5em]{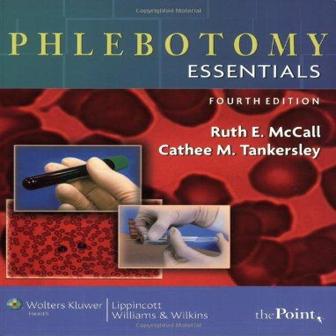}} & Who wrote this book? & A. John D. Smith \newline \textcolor{orange}{B. Ruth E. McCall BS  MT(ASCP)} \newline C. Michael A. Johnson \newline D. Emily J. Brown \\
OCRVQA & \raisebox{-\totalheight}{\includegraphics[height=5em]{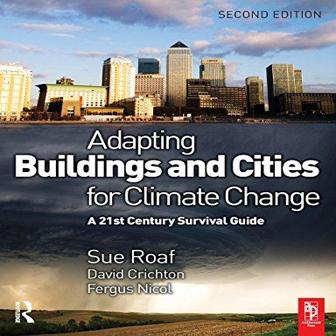}} & What is the title of this book? & A. Climate Change and Urban Adaptation Strategies \newline B. Building a Sustainable Future: Climate Change Adaptation \newline C. Adapting Urban Spaces: Sustainability in the 21st Century \newline \textcolor{orange}{D. Adapting Buildings and Cities for Climate Change} \\
OCRVQA & \raisebox{-\totalheight}{\includegraphics[height=5em]{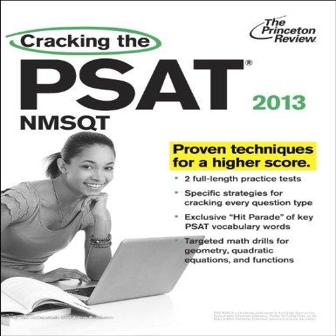}} & What is the title of this book? & \textcolor{orange}{A. Cracking the PSAT/NMSQT, 2013 Edition (College Test Preparation)} \newline B. Cracking the SAT, 2013 Edition (College Test Preparation) \newline C. Crushing the PSAT/NMSQT, 2013 Edition \newline D. Cracking the PSAT, 2014 Edition (College Test Preparation) \\

    \bottomrule
    \end{tabular}
    \caption{\textbf{Examples of \textit{VMCBench} (4/7).} Each example has a dataset source, image, question, and four choices (correct choice highlighted in \textcolor{orange}{orange}).}
    \label{tab:examples5}
\end{table*}

\begin{table*}[!tb]

\rowcolors{2}{white}{light-light-gray}
\setlength\tabcolsep{6pt}
\renewcommand{\arraystretch}{1.1}
\small
\centering

    \begin{tabular}{p{0.1\linewidth}p{0.1\linewidth}p{0.3\linewidth}p{0.4\linewidth}}
    \toprule
    \textbf{Source} & \textbf{Image} & \textbf{Question} & \textbf{Choices} \\
    \midrule

OKVQA & \raisebox{-\totalheight}{\includegraphics[height=5em]{images/8935.jpg}} & What food is being sold? & A. hamburger \newline B. sandwich \newline \textcolor{orange}{C. hot dog} \newline D. kebab \\
OKVQA & \raisebox{-\totalheight}{\includegraphics[height=5em]{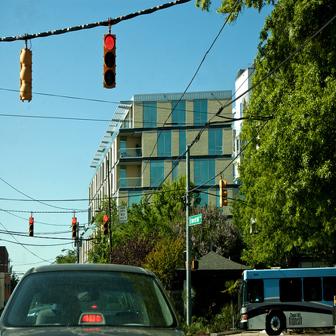}} & What is the proper response when traveling in a vehicle and seeing a red traffic light? & A. proceed with caution \newline B. speed up to clear the intersection \newline \textcolor{orange}{C. stop} \newline D. yield only if necessary \\
OKVQA & \raisebox{-\totalheight}{\includegraphics[height=5em]{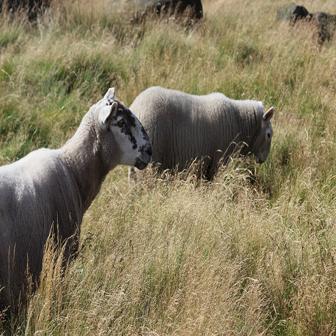}} & How long does this animal usually live? & A. 25 years \newline \textcolor{orange}{B. 10 years} \newline C. 20 years \newline D. 8 years \\
RealWorldQA & \raisebox{-\totalheight}{\includegraphics[height=5em]{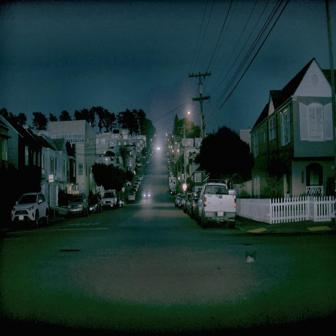}} & How many oncoming vehicles are there? & A. 4 \newline B. 3 \newline C. 1 \newline \textcolor{orange}{D. 2} \\
RealWorldQA & \raisebox{-\totalheight}{\includegraphics[height=5em]{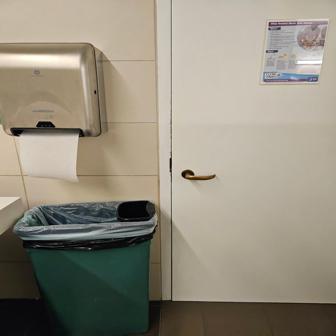}} & Which way does this door open? & A. The door opens outward, swinging to the right. \newline B. The door is a sliding door, moving to the right. \newline \textcolor{orange}{C. The door opens inward, swinging to the right.} \newline D. The door opens outward, swinging to the left. \\
RealWorldQA & \raisebox{-\totalheight}{\includegraphics[height=5em]{images/4311.jpg}} & Which object is bigger than the other? & \textcolor{orange}{A. Both objects are the same size.} \newline B. The right object is bigger. \newline C. The left object has more volume. \newline D. The left object is bigger. \\
SEEDBench & \raisebox{-\totalheight}{\includegraphics[height=5em]{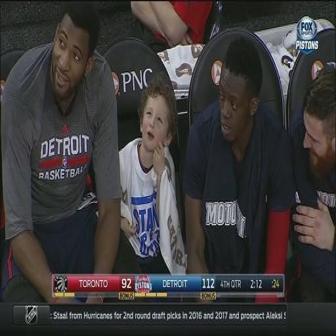}} & What can be found in the image? & \textcolor{orange}{A. A group of people sitting down and a young boy with a basketball player.} \newline B. A basketball player signing autographs for a line of fans. \newline C. A group of musicians playing instruments \newline D. Several students in a classroom setting \\
SEEDBench & \raisebox{-\totalheight}{\includegraphics[height=5em]{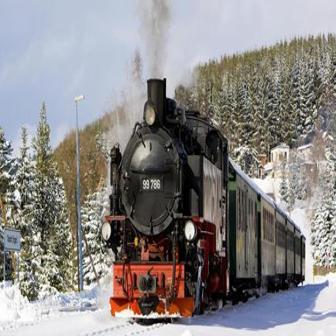}} & Which object is emitting smoke in the image? & A. Factory smokestack in the background \newline B. House chimney \newline C. Chimney of a nearby house \newline \textcolor{orange}{D. Train} \\
SEEDBench & \raisebox{-\totalheight}{\includegraphics[height=5em]{images/2614.jpg}} & What is the boy doing in the image? & A. Jumping \newline \textcolor{orange}{B. Smiling} \newline C. Reading a book \newline D. Waving \\

    \bottomrule
    \end{tabular}
    \caption{\textbf{Examples of \textit{VMCBench} (5/7).} Each example has a dataset source, image, question, and four choices (correct choice highlighted in \textcolor{orange}{orange}).}
    \label{tab:examples7}
\end{table*}

\begin{table*}[!tb]

\rowcolors{2}{white}{light-light-gray}
\setlength\tabcolsep{6pt}
\renewcommand{\arraystretch}{1.1}
\small
\centering

    \begin{tabular}{p{0.1\linewidth}p{0.1\linewidth}p{0.3\linewidth}p{0.4\linewidth}}
    \toprule
    \textbf{Source} & \textbf{Image} & \textbf{Question} & \textbf{Choices} \\
    \midrule

ScienceQA & \raisebox{-\totalheight}{\includegraphics[height=5em]{images/2139.jpg}} & What is the name of the colony shown? & \textcolor{orange}{A. Rhode Island} \newline B. Connecticut \newline C. New Hampshire \newline D. Massachusetts \\
ScienceQA & \raisebox{-\totalheight}{\includegraphics[height=5em]{images/2001.jpg}} & Which continent is highlighted? & A. Australia \newline B. Arctic \newline C. South America \newline \textcolor{orange}{D. Antarctica} \\
ScienceQA & \raisebox{-\totalheight}{\includegraphics[height=5em]{images/1815.jpg}} & What can Turner and Mona trade to each get what they want? & \textcolor{orange}{A. Turner can trade his tomatoes for Mona's broccoli.} \newline B. Turner can trade his oranges for Mona's water. \newline C. Turner can trade his water for Mona's almonds. \newline D. Turner can trade his sandwich for Mona's hot dog. \\
TableVQA-Bench & \raisebox{-\totalheight}{\includegraphics[height=5em]{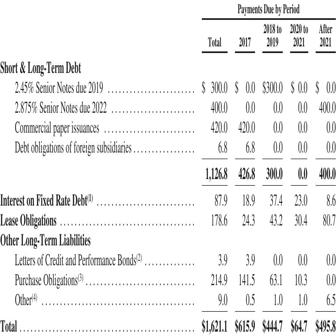}} & What is the total amount of Purchase Obligations due after 2021? & A. \$6.5 \newline B. \$80.7 \newline \textcolor{orange}{C. \$0.0} \newline D. \$214.9 \\
TableVQA-Bench & \raisebox{-\totalheight}{\includegraphics[height=5em]{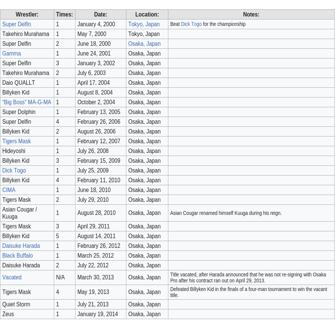}} & what other name did asian cougar have? & A. Gamma \newline B. Kooga \newline \textcolor{orange}{C. Kuuga} \newline D. Black Buffalo \\
TableVQA-Bench & \raisebox{-\totalheight}{\includegraphics[height=5em]{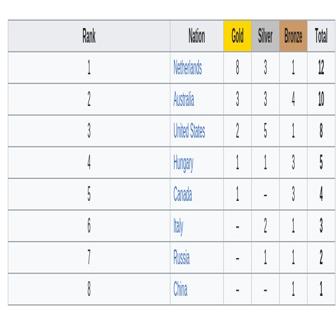}} & how many metals did netherlands and the us win together? & A. 18 \newline \textcolor{orange}{B. 20} \newline C. 22 \newline D. 19 \\
TextVQA & \raisebox{-\totalheight}{\includegraphics[height=5em]{images/4635.jpg}} & what number is shown? & A. 21 \newline B. 25 \newline \textcolor{orange}{C. 22} \newline D. 20 \\
TextVQA & \raisebox{-\totalheight}{\includegraphics[height=5em]{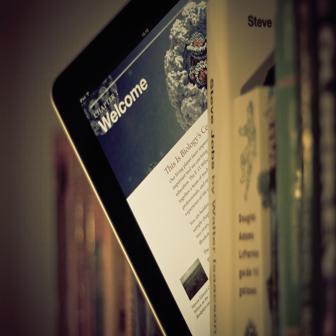}} & what male name is written on the white book? & A. mike \newline B. sean \newline C. dave \newline \textcolor{orange}{D. steve} \\
TextVQA & \raisebox{-\totalheight}{\includegraphics[height=5em]{images/4865.jpg}} & which company is giving the presentation? & A. ibm \newline B. sap \newline C. google \newline \textcolor{orange}{D. oracle} \\

    \bottomrule
    \end{tabular}
    \caption{\textbf{Examples of \textit{VMCBench} (6/7).} Each example has a dataset source, image, question, and four choices (correct choice highlighted in \textcolor{orange}{orange}).}
    \label{tab:examples8}
\end{table*}

\begin{table*}[!tb]

\rowcolors{2}{white}{light-light-gray}
\setlength\tabcolsep{6pt}
\renewcommand{\arraystretch}{1.1}
\small
\centering

    \begin{tabular}{p{0.1\linewidth}p{0.1\linewidth}p{0.3\linewidth}p{0.4\linewidth}}
    \toprule
    \textbf{Source} & \textbf{Image} & \textbf{Question} & \textbf{Choices} \\
    \midrule

VQAv2 & \raisebox{-\totalheight}{\includegraphics[height=5em]{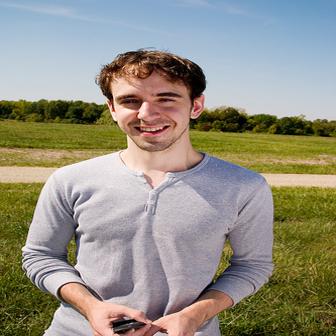}} & How many buttons on the man's shirt? & A. 5 \newline B. 2 \newline \textcolor{orange}{C. 3} \newline D. 1 \\
VQAv2 & \raisebox{-\totalheight}{\includegraphics[height=5em]{images/9364.jpg}} & How many men are in the picture? & A. 2 \newline \textcolor{orange}{B. 1} \newline C. 0 \newline D. 3 \\
VQAv2 & \raisebox{-\totalheight}{\includegraphics[height=5em]{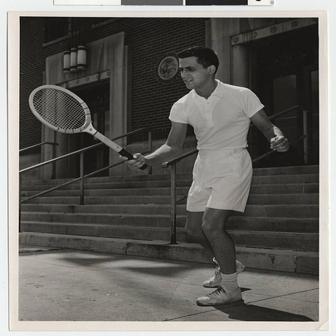}} & What type of shoes is the man wearing? & \textcolor{orange}{A. tennis shoes} \newline B. dress shoes \newline C. hiking boots \newline D. sandals \\
VizWiz & \raisebox{-\totalheight}{\includegraphics[height=5em]{images/6886.jpg}} & What is this a picture of? & A. squirrel \newline B. wolf \newline C. hamster \newline \textcolor{orange}{D. dog} \\
VizWiz & \raisebox{-\totalheight}{\includegraphics[height=5em]{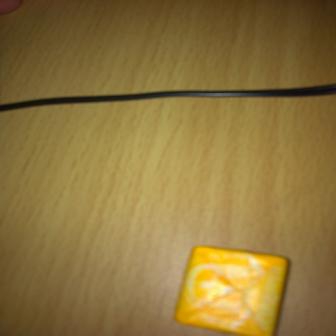}} & What color is this Starburst?  & \textcolor{orange}{A. yellow} \newline B. blue \newline C. green \newline D. red \\
VizWiz & \raisebox{-\totalheight}{\includegraphics[height=5em]{images/6891.jpg}} & What is this? It's from Schwans. & \textcolor{orange}{A. hawaiian style pizza} \newline B. cheese pizza \newline C. meat lovers pizza \newline D. vegetable pizza \\

    \bottomrule
    \end{tabular}
    \caption{\textbf{Examples of \textit{VMCBench} (7/7).} Each example has a dataset source, image, question, and four choices (correct choice highlighted in \textcolor{orange}{orange}).}
    \label{tab:examples10}
\end{table*}